%% file: main.tex
\documentclass{article}
\usepackage{davinci_preprint,times}
\usepackage[utf8]{inputenc} 
\usepackage[T1]{fontenc}    
\usepackage[hypertexnames=false]{hyperref}
\usepackage{url}
\usepackage{booktabs}
\usepackage{amsmath,amssymb}
\usepackage{amsfonts}
\usepackage{microtype}
\usepackage[table]{xcolor}
\usepackage{enumitem}
\usepackage{graphicx}
\usepackage{algorithm}
\usepackage{algpseudocode}
\usepackage{multirow}
\usepackage{wrapfig}
\usepackage{needspace}
\usepackage{subcaption}
\usepackage{wrapfig}

\hypersetup{
  pdfauthor={Xiaoyang Yuan, Ji Zhao, Xingyu Hu, Wenhan Wang, Yutao Hu, Yi Bin, Wenqi Shao, Qiaosheng Zhang, Pengfei Liu, Yu Qiao},
  pdftitle={Davinci-Tool: Scaling Agentic Tool-Use Supervision with Verifiable Executable Environments},
  colorlinks=true,
  linkcolor=blue!50!black,
  citecolor=blue!50!black,
  urlcolor=blue!50!black
}

\title{Davinci-Tool: Scaling Agentic Tool-Use Supervision with Verifiable Executable Environments}

\author{%
\parbox[t]{\dimexpr\textwidth-2\tabcolsep\relax}{%
\centering
Xiaoyang~Yuan\textsuperscript{1,2}\thanks{Equal contribution.}\quad
Ji~Zhao\textsuperscript{2,3}\footnotemark[1]\quad
Xingyu~Hu\textsuperscript{2,3}\footnotemark[1]\quad
Wenhan~Wang\textsuperscript{2}\quad
Yutao~Hu\textsuperscript{4}\quad
Qiaosheng~Zhang\textsuperscript{2}\quad
Pengfei~Liu\textsuperscript{2}\quad
Yu~Qiao\textsuperscript{2}\quad
Yi~Bin\textsuperscript{1}\thanks{Corresponding author}\quad
Wenqi~Shao\textsuperscript{2}\thanks{Project lead.}
\endgraf\smallskip
\normalfont\small
\mbox{\textsuperscript{1}Tongji University}\quad
\mbox{\textsuperscript{2}Shanghai Innovation Institute}\quad
\mbox{\textsuperscript{3}University of Science and Technology of China}\quad
\mbox{\textsuperscript{4}Southeast University}
}%
}

\date{}

\begin{document}

\maketitle

\begin{abstract}
Reliable agentic tool use requires training data that pair diverse multi-step workflows with executable evidence that the requested outcomes were achieved. Scaling either trajectory coverage or verification alone is insufficient as new environments introduce distinct states, rules, and dependencies that must also be reflected in the success criteria. To address this challenge, we introduce \textbf{Davinci-Tool}, a $30$B tool-use model, and \textbf{Davinci-ToolData}, an execution-grounded dataset designed to scale environment coverage and outcome verification together. Davinci-ToolData preserves restorable states, execution traces, outcomes, and verifiers as a unified substrate for supervision. Two complementary pipelines populate this substrate: {Execution-Grounded Task Synthesis (EGTS)} derives verified demonstrations from independently authored systems, and {Executable Environment Synthesis (EES)} synthesizes auditable environments, state-grounded tasks, and executable verifiers across single and coordinated domains. These pipelines yield $30{,}395$ verified trajectories spanning $2{,}260$ invoked tool names. We first train on these demonstrations with supervised fine-tuning, then reuse EES states and verifiers to compare alternative actions and construct outcome-grounded preferences for Direct Preference Optimization. Davinci-Tool improves Qwen3-Coder-30B-A3B-Instruct from $52.4$ to $88.2$ on $\tau^2$-bench, from $30.4$ to $62.8$ on BFCL v4 multi-turn, and from $16.9$ to $56.2$ on MCP-Atlas, outperforming Qwen3-Coder-480B-A35B-Instruct on all three benchmarks with only $1/16$ of its total parameters. \textit{Code is available at \url{https://github.com/MassLab-SII/Davinci-Tool}.}
\end{abstract}

\begin{figure}[htbp]
\centering
\includegraphics[width=0.88\textwidth]{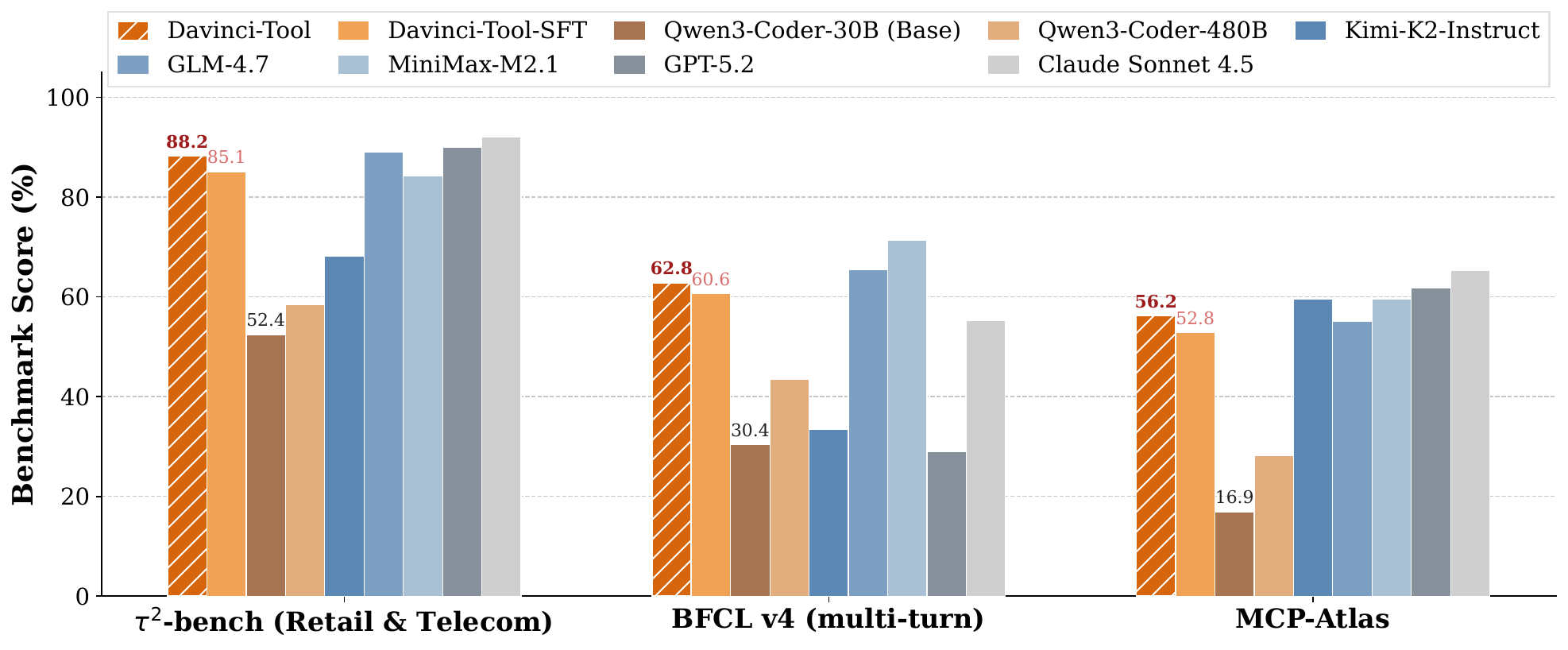}
\caption{Performance on $\tau^2$-bench, BFCL v4 multi-turn, and MCP-Atlas. Davinci-Tool-30B after DPO is shown with hatching, with its SFT checkpoint included for comparison. Detailed results and score provenance are reported in Table~\ref{tab:main}.}
\label{fig:benchmark-comparison}
\end{figure}

\input{sections/introduction}
\input{sections/related_work}

\input{sections/sft_data_engine}

\input{sections/dpo_data_engine}
\input{sections/experiments}
\input{sections/conclusion}

\bibliographystyle{davinci_preprint}
\bibliography{references}

\clearpage
\appendix
\input{sections/appendix}

\end{document}

%% file: sections/introduction.tex
\section{Introduction}

Tool use enables language models to move beyond generating responses and perform real-world tasks, powering applications such as coding agents~\citep{yang2024sweagent} and deep research systems~\citep{zheng2025deepresearcher,li2025webthinker}. Such tasks require stateful multi-step interaction, where models select tools, interpret observations, and adapt actions toward desired outcomes~\citep{zhou2023webarena,xie2024osworld,patil2025bfcl,yao2024taubench}. Training such behavior therefore requires data that connect user intent, multi-step execution, and verifiable environment outcomes.

However, scaling such data presents two coupled challenges: covering diverse tool-use workflows and verifying their outcomes. Prior efforts improve either workflow coverage through expanded tool inventories and synthetic interactions~\citep{qin2023toolllm,liu2024toolace,wang2025mcpflow} or reliability through execution validation and executable verification~\citep{liu2024toolace,ouajdi2026executionfirst,song2026envscaler}. As environments become more diverse and cross-domain, verification must capture environment-specific states, rules, and dependencies. Therefore, environment coverage and outcome verification need to scale together.

Environment expansion follows two complementary paradigms. Independently authored systems provide realistic tools and operational rules but require adaptation of reset, execution, and outcome-checking interfaces, as exemplified by ToolSandbox and AppWorld~\citep{lu2024toolsandbox,trivedi2024appworld}. Synthesized environments provide controllable states and executable criteria for systematic task generation but may inherit assumptions from the environment generator, as explored by EnvScaler and AutoWebWorld~\citep{song2026envscaler,wu2026autowebworld}. These complementary sources motivate a unified execution-grounded framework that supports heterogeneous environments under shared outcome verification.

To address these settings, we introduce \textbf{Davinci-Tool}, a model for reliable multi-step tool use in stateful environments, together with \textbf{Davinci-ToolData}, an execution-grounded training dataset spanning diverse tool-use environments. Our central design principle is to preserve executable evidence throughout the training pipeline: each example is grounded in a restorable initial state, an action--observation trace, the resulting environment outcome, and an executable verification result. This common interface allows data from heterogeneous environment sources to be accepted under a shared outcome-based criterion and enables the same states and verifiers to support both supervised and preference learning.

To construct \textbf{Davinci-ToolData}, we combine independently authored systems with synthesized executable environments. For authored systems, \textbf{Execution-Grounded Task Synthesis (EGTS)} explores supported interactions to establish achievable outcomes and their evidence, synthesizes user tasks from these outcomes, and replays them from corresponding initial states to collect verified trajectories. This separates outcome discovery from demonstration construction: exploration determines what can be achieved, while replay collects trajectories conditioned on the synthesized user request. For synthesized environments, \textbf{Executable Environment Synthesis (EES)} constructs and audits rule-governed domains before generating state-grounded tasks and executable verifiers. EES produces scenarios within individual domains (\textbf{EES-Single}) and composes compatible domains through shared entities, policies, and state dependencies before generating coordinated tasks over the resulting executable world (\textbf{EES-Cross}). Together, these strategies provide complementary supervision from single-domain to cross-domain workflows. Davinci-ToolData contains $30{,}395$ verified trajectories covering $2{,}260$ distinct invoked tool names (Table~\ref{tab:corpus-coverage}).

We first fine-tune Qwen3-Coder-30B-A3B-Instruct~\citep{qwen2025qwen3} on Davinci-ToolData, producing Davinci-Tool-30B-SFT. To complement successful demonstrations with comparisons between alternative decisions, we reuse restorable EES scenarios and their executable verifiers for decision-level preference learning. At selected points in SFT rollouts, we restore the same executable context, sample alternative decisions, and use verified downstream outcomes to construct preference pairs. Only divergent assistant turns are optimized with Direct Preference Optimization (DPO)~\citep{rafailov2023dpo}. Unlike approaches that rely on tree search with intermediate self-evaluation~\citep{xie2024mctsdpo,putta2024agentq}, our construction uses bounded sibling rollouts and executable outcome checks without an additional expert action generator or process evaluator.

Davinci-Tool-30B achieves substantial gains across three held-out tool-use benchmarks. SFT improves the base model from $52.4$ to $85.1$ on $\tau^2$-bench, from $30.4$ to $60.6$ on BFCL v4 multi-turn, and from $16.9$ to $52.8$ on MCP-Atlas, while decision-level DPO further raises these scores to $88.2$, $62.8$, and $56.2$, respectively. Despite using only $1/16$ as many total parameters, Davinci-Tool-30B outperforms Qwen3-Coder-480B-A35B-Instruct~\citep{qwen2025qwen3} on all three benchmarks. Source-composition studies further show that the full EGTS, EES-Single, and EES-Cross mixture provides the strongest SFT performance among evaluated configurations, while outcome-grounded preference learning provides additional gains after SFT.

Our contributions are:
\begin{itemize}[leftmargin=1.5em,itemsep=1pt,topsep=2pt,parsep=0pt]

    \item We introduce \textbf{Davinci-Tool}, a tool-use model trained through an executable-environment framework that connects verified demonstration construction with outcome-grounded, decision-level preference learning.

    \item We construct \textbf{Davinci-ToolData}, an execution-grounded dataset with $30{,}395$ verified tool-use demonstrations and $2{,}260$ invoked tool names. EGTS builds task-conditioned demonstrations through exploration and verified replay, while EES constructs and audits executable environments for single-domain and coordinated cross-domain tasks.

    \item We demonstrate strong tool-use performance at the $30$B scale. Davinci-Tool-30B improves over its backbone by $35.8$, $32.4$, and $39.3$ percentage points on $\tau^2$-bench, BFCL v4 multi-turn, and MCP-Atlas, respectively, outperforming Qwen3-Coder-480B-A35B-Instruct across all three benchmarks. The same Davinci-ToolData supervision also substantially improves the tested smaller models.

\end{itemize}

%% file: sections/related_work.tex
\section{Related Work}
\label{sec:related}

\textbf{Tool-use data construction.}
ToolLLM~\citep{qin2023toolllm} and ToolACE~\citep{liu2024toolace} scale supervision through synthesis, while MCP-Flow~\citep{wang2025mcpflow} and TOUCAN~\citep{xu2025toucan} collect trajectories from MCP environments. Execution-first methods such as SyntheticAgentTraceQA~\citep{ouajdi2026executionfirst} and Trajectory2Task~\citep{wang2026trajectory2task} derive tasks from observed executions. Trajectory2Task additionally collects successful task rollouts, while AgentEvolver~\citep{zhai2025agentevolver} uses exploration to collect RL data. EGTS builds on this direction for heterogeneous authored systems: an originating execution provides outcome evidence and replay guidance, and the synthesized request is re-executed from the corresponding initial state to collect a verified task-conditioned demonstration. Unlike AgentTrek's tutorial-guided replay~\citep{xu2025agenttrek}, EGTS derives both guidance and verification evidence from the originating execution.

\textbf{Environment synthesis.}
EnvScaler~\citep{song2026envscaler} constructs executable environments, task scenarios, and validation functions, while ASTRA~\citep{tian2026astra} synthesizes verifiable sub-environments around question--answer decompositions. EES follows this domain-first direction by auditing reusable executable domains before task generation. EES-Single generates state-grounded tasks and verifiers within individual domains, while EES-Cross composes audited domains through shared entities, roles, policies, and state dependencies before coordinated task generation. AppWorld~\citep{trivedi2024appworld} and Toolathlon~\citep{li2026toolathlon} support multi-application interaction. In contrast, EES-Cross uses executable domain composition to construct verified training supervision.

\textbf{Preference learning.}
ETO~\citep{song2024eto} learns from success--failure trajectory pairs, while MCTS-DPO~\citep{xie2024mctsdpo}, Agent Q~\citep{putta2024agentq}, and AlphaLLM-CPL~\citep{wang2024alphallmcpl} derive finer-grained preferences through search-based alternatives and evaluation. Our construction branches from restorable EES states: the SFT policy generates bounded sibling decisions and continuations under the same executable context, and verified outcomes label the divergent turns. This procedure uses bounded sibling rollouts rather than recursive tree search and executable verification rather than a separate process evaluator. It also reuses the restoration and verification infrastructure from demonstration construction. Further discussion appears in Appendix~\ref{app:additional-related}.

%% file: sections/sft_data_engine.tex
\section{Verified Demonstration Construction}
\label{sec:data-engine}
\label{sec:overview}
\label{sec:evidence-contract}

\begin{figure}[htbp]
\centering
\includegraphics[width=1.0\textwidth]{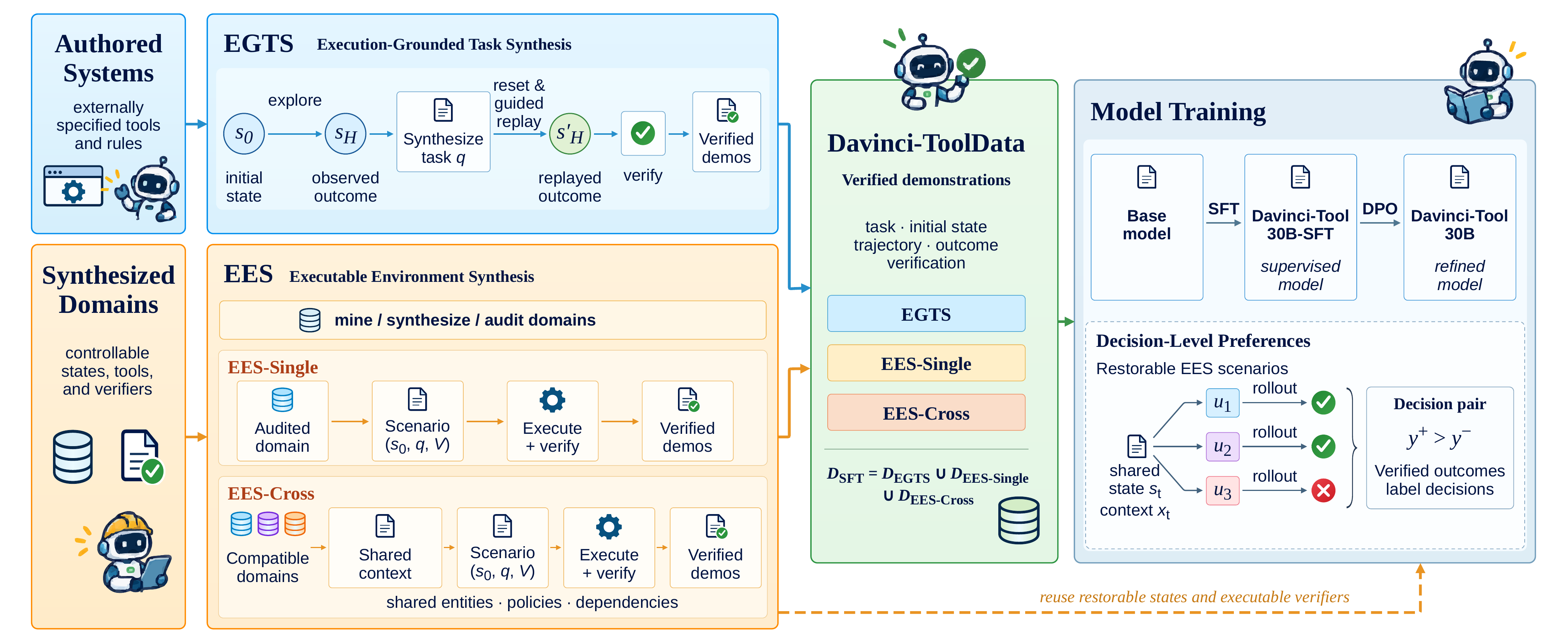}
\vspace{-1em}
\caption{Construction of Davinci-ToolData and training of Davinci-Tool. EGTS converts explored outcomes in authored systems into verified task-conditioned replay demonstrations, while EES constructs audited executable environments for verified single-domain and cross-domain trajectories. The resulting demonstrations support SFT, and restorable EES states and verifiers are subsequently reused for decision-level DPO.}
\label{fig:tool-use-overview}
\vspace{-1em}
\end{figure}

To jointly scale environment coverage and outcome verification, Davinci-ToolData follows two complementary construction routes (Figure~\ref{fig:tool-use-overview}). Both operate under a shared execution and verification interface but differ in how their environments and supervision are obtained. For independently authored systems, EGTS identifies feasible outcomes through exploration, synthesizes outcome-grounded tasks, and collects verified task-conditioned demonstrations through guided replay. For synthesized environments, EES constructs and audits reusable executable domains, then generates state-grounded tasks and executable verifiers for individual domains (EES-Single) and coordinated cross-domain workflows (EES-Cross) linked by shared entities, policies, and state dependencies.

\textbf{Definition of tool-use trajectory.}
We represent a resettable tool environment as
$W=(\mathcal{S},\mathcal{T},\rho,P,\mathcal{E})$,
where $\mathcal{S}$ is the state space, $\mathcal{T}$ is the available tool set, $\rho(s_0)$ restores the environment to an initial-state snapshot $s_0$, $P$ executes tool calls and produces state transitions and observations, and $\mathcal{E}$ is the outcome-evidence space. The restoration operator $\rho$ enables replaying tasks or branching alternative decisions from the same executable state.

Given a user task $q$ and initial state $s_0$, the model maintains an interaction history $h_t=(a_1,o_1,\ldots,a_{t-1},o_{t-1})$. At step $t$, the tool-use policy $\pi_\theta$ selects a tool call $a_t$ and receives observation $o_t$:
$
a_t \sim \pi_\theta(\cdot\mid q,h_t,\mathcal{T}),
\,
(s_t,o_t)=P(s_{t-1},a_t).
$
After $H$ tool interactions, the model produces a final response $y$, yielding
$
\tau=(a_1,o_1,\ldots,a_H,o_H,y)
$
with terminal state $s_H$. Given outcome evidence $\xi\in\mathcal{E}$, an execution is accepted when
$
\operatorname{Verify}_W(q,s_0,\tau,s_H,\xi)=1.
$
EGTS evaluates replay using exploration-derived evidence, whereas EES instantiates verification through executable task predicates. Accepted trajectories provide SFT supervision, while restorable EES states and verifiers further support decision-level preference learning (Section~\ref{sec:preference-construction}). Appendix~\ref{app:shared-evidence} details the shared execution and verification interface.

\subsection{EGTS: From Explored Outcomes to Verified Demonstrations}
\label{sec:egts}

EGTS separates \emph{outcome discovery} from \emph{task-conditioned demonstration construction}. An exploratory execution first establishes a feasible outcome in an independently authored environment $W$. EGTS then synthesizes a user task $q$ from captured outcome evidence, restores the same initial state with $\rho(s_0)$, and collects a new trajectory conditioned on $q$. Thus, exploration determines \emph{what can be achieved}, while replay demonstrates \emph{how to achieve it for the user request}. This design builds on execution-first task synthesis~\citep{ouajdi2026executionfirst,wang2026trajectory2task} while separating discovery traces from SFT supervision.

\textbf{Outcome exploration and evidence collection.}
EGTS adapts each authored system to the shared interface, providing callable tools $\mathcal{T}$, restoration $\rho$, transitions $P$, and source-specific evidence capture in $\mathcal{E}$. Starting from $s_0$, coverage-guided exploration produces trajectory $\tau_{\mathrm{exp}}$ and terminal state $s_H$. The collector extracts evidence $\xi_{\mathrm{exp}}\in\mathcal{E}$ characterizing the achieved outcome:
\vspace{-0.1in}
\begin{equation}
(s_0,\tau_{\mathrm{exp}},s_H)
\xrightarrow{\text{capture in }W}
\xi_{\mathrm{exp}}.
\label{eq:egts-exploration}
\end{equation}
For state-modifying environments, $\xi_{\mathrm{exp}}$ retains relevant state transitions, whereas for retrieval-oriented environments, it retains grounded observations. Failed executions, insufficient evidence, or outcomes without usable evidence are discarded. Reusable skill guidance and prior execution experience help exploration reach more involved workflows~\citep{yang2025exif,ramrakhya2025autoplay}.

\textbf{Task synthesis.}
Only after an outcome is executed and recorded does EGTS generate a user task,
$q \sim G_{\mathrm{task}}(s_0,\tau_{\mathrm{exp}},s_H,\xi_{\mathrm{exp}})$.
The task specifies the desired outcome without exposing the exploratory action sequence. For mutation tasks, it requests state changes supported by the observed transition, whereas for retrieval tasks, it requests information grounded in $\xi_{\mathrm{exp}}$. Candidate tasks are retained only when they are specific, natural, feasible from $s_0$, and fully supported by evidence.

\textbf{Trajectory construction and verification.}
EGTS extracts task-relevant procedural information from $(\tau_{\mathrm{exp}},\xi_{\mathrm{exp}})$ as replay guide $g$, invokes $\rho(s_0)$, and executes task $q$ to obtain trajectory $\tau_{\mathrm{rep}}$ terminating in $s'_H$. The guide is used only during collection and excluded from serialized SFT examples.

The replay is accepted when
$\operatorname{Verify}_W(q,s_0,\tau_{\mathrm{rep}},s'_H,\xi_{\mathrm{exp}})=1$.
For mutation tasks, this requires recovering the canonical outcome from exploration, including
$\phi_W(s'_H)=\phi_W(s_H)$,
where $\phi_W$ removes source-specific volatile state irrelevant to task semantics. For retrieval tasks, tool observations and final responses must be grounded in $\xi_{\mathrm{exp}}$. Additional checks cover tool-call executability, argument--observation consistency, action ordering, and task satisfaction.

Each accepted SFT example retains user task $q$, visible tool set $\mathcal{T}$, and task-conditioned replay trajectory $\tau_{\mathrm{rep}}$, together with verification provenance. The exploratory trajectory $\tau_{\mathrm{exp}}$ and replay guide $g$ support construction but are not used as model-visible SFT context. Appendix~\ref{app:egts-details} provides the full exploration, replay, and verification procedure.

\subsection{EES: From Executable Domains to Verified Trajectories}
\label{sec:ees}

Unlike EGTS, which adapts authored systems and derives tasks from explored outcomes, EES jointly constructs executable environments, state-grounded tasks, and their success criteria. It first synthesizes and audits reusable domains, then instantiates verified scenarios within individual domains (EES-Single) or composed domains (EES-Cross). Since EES controls state initialization, transitions, and verification, each accepted trajectory retains a restorable state and executable predicates that can be reused for preference construction (Section~\ref{sec:preference-construction}).

\subsubsection{EES-Single: Audited Single-Domain Scenarios}
\label{sec:pes-single}
\label{sec:programmatic-environments}

\textbf{Domain synthesis and audit.}
An EES domain is $d_i=(W_i,\Gamma_i)$, where
$W_i=(\mathcal{S}_i,\mathcal{T}_i,\rho_i,P_i,\mathcal{E}_i)$
follows the shared tool-environment interface and $\Gamma_i$ specifies entity schemas and operational rules. Staged synthesis produces schemas, tools, and transition logic. An executor--auditor loop exercises $\mathcal{T}_i$ over initialized states and checks arguments, observations, state transitions, and compliance with $\Gamma_i$. Detected inconsistencies trigger revisions before task generation~\citep{song2026envscaler}.

\textbf{Scenario and verifier construction.}
For an audited domain, EES generates a scenario $(s_0,q,\mathcal{V})$, where $s_0\in\mathcal{S}_i$, $q$ is grounded in $s_0$, and $\mathcal{V}=\{v_1,\ldots,v_J\}$ contains executable predicates for atomic requirements. A mutation target already satisfied in $s_0$ is regenerated so that success requires an interaction-induced state change. Executing $q$ from $\rho_i(s_0)$ produces trajectory $\tau$ and terminal state $s_H$. Its predicate evaluations form scenario evidence $\xi_{\mathcal V}\in\mathcal{E}_i$, and the trajectory is accepted when
\vspace{-0.09in}
\begin{equation}
\operatorname{Verify}_{W_i}(q,s_0,\tau,s_H,\xi_{\mathcal V})
=\prod_{j=1}^{J}\mathbb{I}[v_j(\tau,s_H)=1]=1.
\label{eq:ees-single-verification}
\end{equation}
%
Domain auditing ensures that the environment follows its contract, while scenario predicates verify completion of the requested task.

\subsubsection{EES-Cross: Composing Domains into Coordinated Workflows}
\label{sec:pes-cross}

EES-Cross composes audited domains before task generation to create executable shared worlds where tasks depend on explicit entity, state, or policy relationships rather than unrelated tool collections.

\textbf{Domain selection and composition.}
Embedding retrieval proposes related domains, an affinity rubric evaluates functional compatibility and potential dependencies, and bounded random walks select connected groups. Namespace maps separate tool names, entity types, identifiers, and state attributes before composition. The resulting
$W_{\mathrm{cross}}=(\mathcal{S}_{\mathrm{cross}},\mathcal{T}_{\mathrm{cross}},\rho_{\mathrm{cross}},P_{\mathrm{cross}},\mathcal{E}_{\mathrm{cross}})$
preserves each domain's local transitions under a unified runtime. Static checks validate initialization, schemas, and namespace consistency.

\textbf{Shared-world construction.}
EES-Cross defines shared entities, user roles, linking fields, cross-domain policies, and coherence constraints over $W_{\mathrm{cross}}$. These relations are checked against domain contracts before $\rho_{\mathrm{cross}}(s_0)$ initializes linked records, making the operational dependencies explicit.

\textbf{Task construction and verification.}
EES-Cross generates task $q$ and predicates $\mathcal{V}_{\mathrm{cross}}$ from the shared context and $s_0$. Requirements may transfer an observation or identifier across domains, enforce operation ordering, or constrain state changes through cross-domain policies. Candidate trajectories first pass runtime checks and are then accepted through the shared interface:
\vspace{-0.1in}
\begin{equation}
    \operatorname{Verify}_{W_{\mathrm{cross}}}
(q,s_0,\tau,s_H,\xi_{\mathcal V_{\mathrm{cross}}})=1,
\end{equation}
where $\xi_{\mathcal V_{\mathrm{cross}}}\in\mathcal{E}_{\mathrm{cross}}$ records cross-domain predicate outcomes. A retained example uses content from a ``Cryo-EM Vendor Call'' note to update a mobile reminder. The agent verifies reminder ownership before the update and separately reports the user's notification preference across the \texttt{personal\_} and \texttt{mobile\_} interfaces (Appendix~\ref{app:cross-trajectory-example}). Appendix~\ref{app:programmatic-details} details EES construction and auditing.

\begin{wraptable}[8]{r}{0.56\textwidth}
\vspace{-1.3em}
\centering
\caption{Verified SFT corpus. Token and call counts are per-trajectory means. Tool counts represent distinct invoked serialized names.}
\label{tab:corpus-coverage}
\vspace{0.2em}
\scriptsize
\setlength{\tabcolsep}{3.5pt}
\begin{tabular}{@{}lrrrr@{}}
\toprule
\textbf{Source} & \textbf{Traj.} & \textbf{Tokens} & \textbf{Calls} & \textbf{Tools} \\
\midrule
EGTS & 13,824 & 10,063.9 & 6.04 & 350 \\
EES-Single & 8,026 & 10,372.7 & 5.87 & 1,610 \\
EES-Cross & 8,545 & 20,444.7 & 7.01 & 1,683 \\
\midrule
Full SFT corpus & 30,395 & 13,063.8 & 6.26 & 2,260 \\
\bottomrule
\end{tabular}
\vspace{-0.5em}
\end{wraptable}

\subsection{Environment Coverage and Corpus Statistics}
\label{sec:corpus-statistics}

\textbf{Environment coverage.}
EGTS covers $8$ authored environment families spanning services, APIs, workspaces, web interaction, and simulation. EES comprises $191$ audited domains across $19$ categories, of which $188$ appear in $246$ retained cross-domain compositions. Appendix~\ref{app:worlds} provides the full source and domain taxonomy.

\begin{figure}[htbp]
\centering
\includegraphics[width=1.0\textwidth]{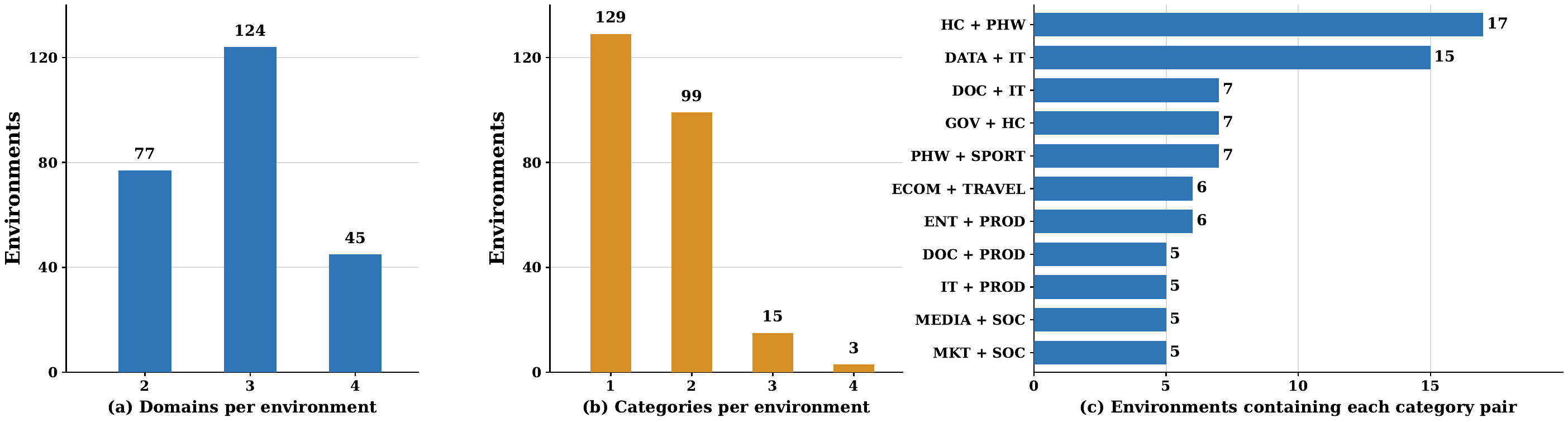}
\vspace{-0.1in}
\caption{Structure of the $246$ retained EES-Cross compositions. Panels (a), (b), and (c) show the number of constituent domains per environment, the number of application categories per environment, and category pairs appearing in at least $5$ compositions, respectively. Pair counts may overlap when an environment spans more than two categories. Category codes are defined in Table~\ref{tab:ees-domain-categories}.}
\label{fig:ees-composition-patterns}
\end{figure}

\textbf{Composition structure.}
Figure~\ref{fig:ees-composition-patterns} shows $77$ two-domain, $124$ three-domain, and $45$ four-domain environments. Among them, $117$ span multiple application categories and $129$ connect distinct systems within one category, demonstrating diversity in both domain count and category composition.

\textbf{Corpus diversity and interaction scale.}
Table~\ref{tab:corpus-coverage} shows that Davinci-ToolData contains $30{,}395$ verified trajectories with $28{,}985$ distinct normalized requests and $436$ declared-tool inventories. Across these inventories, $4{,}672$ tool names are exposed and $2{,}260$ are invoked, producing $190{,}410$ tool calls. Source-level tool counts overlap and are therefore not additive. The median serialized length is $9{,}284$ tokens, with $957$ trajectories ($3.15\%$) exceeding $32{,}768$ tokens before training-time handling. EES-Cross has the highest mean length and tool-call count, reflecting additional cross-domain coordination. Appendix~\ref{app:corpus-statistics} provides source-level distributions.

Beyond corpus scale, representation alignment shows source-specific execution-pattern neighborhoods (Section~\ref{app:trajectory-alignment}). Paired base/SFT analysis characterizes benchmark-specific execution changes associated with task recovery (Section~\ref{app:execution-analysis}). Length-controlled analysis shows that EES-Cross introduces domain transitions, entity transfer, and read--write dependencies beyond interaction length (Section~\ref{app:cross-structure-analysis}).

%% file: sections/dpo_data_engine.tex
\section{Outcome-Grounded Model Training}
\label{sec:model}

Davinci-Tool is trained in two stages. We first perform supervised fine-tuning (SFT) on the verified demonstrations from Section~\ref{sec:data-engine}, then reuse restorable EES scenarios and executable verifiers to compare alternative decisions from the same interaction context. Verified downstream outcomes define decision-level preferences for DPO.

\subsection{Supervised Fine-Tuning}
\label{sec:sft}

We fine-tune Qwen3-Coder-30B-A3B-Instruct~\citep{qwen2025qwen3} on Davinci-ToolData's $30{,}395$ verified demonstrations. Each example contains the user task, visible tools, assistant actions, tool observations, and the final response. SFT optimizes assistant tokens conditioned on the preceding interaction, while tool outputs remain in context but are excluded from the loss. The resulting policy $\pi_{\mathrm{SFT}}$ is used for preference construction. Training details are provided in Appendix~\ref{app:training-details}.

\subsection{Outcome-Grounded Preference Construction}
\label{sec:preference-construction}

A verified demonstration provides one successful execution but does not compare alternative decisions from the same interaction state. We construct such comparisons by branching from $\pi_{\mathrm{SFT}}$ rollouts in EES-Single and EES-Cross scenarios. Restorable states allow alternatives to start from the same executable state, while verifiers provide a common outcome criterion for completed branches.

\textbf{Shared decision contexts.}
For each selected scenario, $\pi_{\mathrm{SFT}}$ generates a trunk trajectory. At a retained decision point $t$, we record
\[
x_t=(q,m_{<t},\mathcal{T}_t),
\]
where $x_t$ is the model-visible context containing task $q$, message prefix $m_{<t}$, and visible tools $\mathcal{T}_t$, while $s_t$ is the corresponding restorable environment state. Up to $B$ trunk decision points are retained as forks, with executable state $s_t$ kept separate from model-visible context $x_t$.

\textbf{Sibling generation and outcome verification.}
At each fork, the trunk decision is retained and $K-1$ alternatives are sampled from the same policy:
\[
u_t^{(1)}=u_t^{\mathrm{trunk}},
\qquad
u_t^{(k)}\sim\pi_{\mathrm{SFT}}(\cdot\mid x_t),
\quad k=2,\ldots,K.
\]
Before each candidate, $\rho$ restores the environment to $s_t$ and the same context $x_t$ is reused. Each branch then continues with $\pi_{\mathrm{SFT}}$ under greedy decoding until termination, producing sibling trajectories $\tau_t^{(k)}$ and terminal states $s_H^{(k)}$. Forks occur only along the trunk and are not recursively expanded. Each sibling receives the verified outcome
\[
r_t^{(k)}
=
\operatorname{Verify}_{W}
(q,s_0,\tau_t^{(k)},s_H^{(k)},\xi_{\mathcal V})
\in\{0,1\}.
\]
Because all candidates start from the same fork state and model-visible context and are evaluated under the same greedy continuation policy and verifier, differences in $r_t^{(k)}$ reflect differences in their verified downstream outcomes.

\textbf{Preference-pair construction.}
When two distinct candidates obtain different verified outcomes, $r_t^+>r_t^-$, we retain
\[
(x_t,y^+,y^-),
\qquad
y^+=u_t^+,\quad y^-=u_t^-.
\]
The divergent assistant turns serve as preferred and rejected responses, while their completed continuations determine the labels and are retained with terminal states and verification records as construction provenance. The resulting pairs form $\mathcal{D}_{\mathrm{pref}}$. This procedure yields $1{,}150$ pairs from EES-Single and EES-Cross scenarios after SFT, separate from the $30{,}395$ SFT demonstrations but reusing the same restorable states and verifiers. Appendix~\ref{app:dpo-details} provides collection details and statistics.

\subsection{Decision-Level Preference Optimization}
\label{sec:decision-dpo}

We initialize $\pi_\theta$ from $\pi_{\mathrm{SFT}}$ and use a frozen copy as the reference policy $\pi_{\mathrm{ref}}$. We apply standard DPO~\citep{rafailov2023dpo} to $\mathcal{D}_{\mathrm{pref}}$, using only the divergent preferred and rejected assistant turns in the objective. Their downstream continuations are used solely to determine the verified preference labels. The resulting checkpoint is Davinci-Tool-30B. Appendix~\ref{app:training-details} provides optimization details.

%% file: sections/experiments.tex
\section{Experiments}
\label{sec:experiments}

We evaluate verified-demonstration training, decision-level preference learning, transfer across model scales, comparison with public tool-use supervision, and contributions of the three data-construction routes. We further analyze supervision structure and resulting execution behavior.

\subsection{Experimental Setup}

\textbf{Benchmarks.}
We evaluate three held-out tool-use suites using their official protocols. $\tau^2$-bench~\citep{barres2025tau2bench} evaluates stateful interaction in Retail and Telecom, and we report both domains and their macro average. BFCL v4 multi-turn~\citep{patil2025bfcl} evaluates multi-turn function calling across four categories. MCP-Atlas~\citep{bandi2026mcpatlas} evaluates heterogeneous MCP tool use. We use the $89$ tasks that do not require external API keys. Evaluation queries and states are excluded from data construction, and all scores are from our evaluations.

\textbf{Models and training.}
Our primary model is Qwen3-Coder-30B-A3B-Instruct~\citep{qwen2025qwen3}. DeepSeek-V4-Pro~\citep{deepseekai2026deepseekv4} is used for trajectory generation. Davinci-Tool-30B-SFT is trained on the $30{,}395$-trajectory Davinci-ToolData corpus, and Davinci-Tool-30B is obtained from this checkpoint using the decision-level preferences in Section~\ref{sec:preference-construction}.

\textbf{Comparison settings.}
We compare with Qwen3-Coder-480B-A35B-Instruct~\citep{qwen2025qwen3}, Kimi-K2-Instruct~\citep{kimi2025k2}, MiniMax-M2.1~\citep{minimax2025m21}, GLM-4.7~\citep{zai2025glm47}, GPT-5.2~\citep{openai2025gpt52}, and Claude-Sonnet-4.5~\citep{anthropic2025claudesonnet45}. We also fine-tune Qwen3-8B and Qwen3-14B~\citep{qwen2025qwen3} on Davinci-ToolData, and compare against the public Toucan-1.5M release~\citep{xu2025toucan} under the same Qwen3-Coder-30B backbone and SFT hyperparameters. MCP-Atlas is omitted for Qwen3-8B/14B because interactions exceed their default $40{,}960$-token context window.

\subsection{Main Results}
\label{sec:main-results}

\begin{table}[htbp]
\vspace{-0.5em}
\centering
\caption{
Main results and model transfer.
Reference models are strong open- and closed-source models included for external comparison.
Underlining marks the best reference-model score in each metric, while bold marks the best score within each of the Qwen3-8B, Qwen3-14B, and primary 30B training groups.
Light and dark shading denote Davinci-ToolData SFT and SFT+DPO, respectively.
}
\label{tab:main}
\setlength{\tabcolsep}{3.6pt}
\renewcommand{\arraystretch}{0.94}

\resizebox{\textwidth}{!}{%
\begin{tabular}{@{}llccccccccc@{}}
\toprule
\textbf{Model}
& \textbf{Training}
& \multicolumn{3}{c}{$\boldsymbol{\tau^2}$\textbf{-bench}}
& \multicolumn{5}{c}{\textbf{BFCL v4 multi-turn}}
& \textbf{MCP-Atlas} \\
\cmidrule(lr){3-5}
\cmidrule(lr){6-10}

&
& \textbf{Retail}
& \textbf{Telecom}
& \textbf{Avg.}
& \textbf{Base}
& \textbf{Long-Context}
& \textbf{Miss-Func}
& \textbf{Miss-Param}
& \textbf{Avg.}
& \textbf{Avg.} \\

\midrule
\multicolumn{11}{@{}l}{\textit{Reference models}} \\
\cmidrule(lr){1-11}

Qwen3-Coder-480B-A35B-Instruct
& ---
& 77.5 & 39.5 & 58.5
& 55.5 & 54.5 & 32.0 & 32.0 & 43.5
& 28.1 \\

Kimi-K2-Instruct
& ---
& 70.6 & 65.8 & 68.2
& 42.0 & 38.5 & 23.0 & 30.0 & 33.4
& 59.6 \\

MiniMax-M2.1
& ---
& 81.6 & 87.0 & 84.3
& \underline{76.5}
& \underline{69.0}
& \underline{71.0}
& \underline{69.0}
& \underline{71.4}
& 59.6 \\

GLM-4.7
& ---
& 79.8
& \underline{98.2}
& 89.0
& 69.5 & 67.0 & 66.0 & 59.0 & 65.4
& 55.1 \\

GPT-5.2
& ---
& 81.6
& \underline{98.2}
& 89.9
& 37.0 & 27.0 & 22.5 & 29.0 & 28.9
& 61.8 \\

Claude-Sonnet-4.5
& ---
& \underline{86.0}
& 98.0
& \underline{92.0}
& 55.5 & 54.5 & 60.0 & 51.0 & 55.3
& \underline{65.2} \\

\midrule
\multicolumn{11}{@{}l}{\textit{Transfer to smaller models}} \\
\cmidrule(lr){1-11}

Qwen3-8B
& Base
& 47.4 & 17.5 & 32.5
& 22.5 & 16.0 & 6.5 & 16.5 & 15.4
& --- \\

\rowcolor[rgb]{.93,.95,.97}
\cellcolor{white}
&
Davinci-ToolData SFT
& \textbf{64.9}
& \textbf{84.2}
& \textbf{74.6}
& \textbf{50.5}
& \textbf{23.5}
& \textbf{29.0}
& \textbf{32.5}
& \textbf{33.9}
& --- \\

\addlinespace[1pt]
\midrule

Qwen3-14B
& Base
& 50.0 & 16.7 & 33.4
& 35.0 & 23.0 & 5.5 & 21.0 & 21.1
& --- \\

\rowcolor[rgb]{.93,.95,.97}
\cellcolor{white}
&
Davinci-ToolData SFT
& \textbf{67.5}
& \textbf{86.0}
& \textbf{76.8}
& \textbf{50.0}
& \textbf{34.0}
& \textbf{42.5}
& \textbf{39.0}
& \textbf{41.4}
& --- \\

\midrule
\multicolumn{11}{@{}l}{\textit{Primary backbone}} \\
\cmidrule(lr){1-11}

Qwen3-Coder-30B-A3B-Instruct
& Base
& 68.7 & 36.0 & 52.4
& 43.0 & 36.0 & 17.0 & 25.5 & 30.4
& 16.9 \\

&
Toucan-1.5M SFT
& 65.8 & 89.5 & 77.7
& 52.5 & 46.0 & 35.0 & 34.0 & 41.9
& 22.5 \\

\rowcolor[rgb]{.93,.95,.97}
\cellcolor{white}
&
Davinci-ToolData SFT
& 73.7
& \textbf{96.5}
& 85.1
& 71.5
& 61.0
& \textbf{63.0}
& 47.0
& 60.6
& 52.8 \\

\rowcolor[rgb]{.80,.882,.937}
\cellcolor{white}
Davinci-Tool-30B
& Davinci-ToolData SFT + DPO
& \textbf{79.8}
& \textbf{96.5}
& \textbf{88.2}
& \textbf{72.5}
& \textbf{62.0}
& 61.5
& \textbf{55.0}
& \textbf{62.8}
& \textbf{56.2} \\

\bottomrule
\end{tabular}%
}
\end{table}

\textbf{Training-stage gains.}
Verified-demonstration SFT establishes most of the improvement on the primary 30B backbone, raising $\tau^2$-bench, BFCL v4, and MCP-Atlas from $52.4/30.4/16.9$ to $85.1/60.6/52.8$. Gains occur in both $\tau^2$ domains and all four BFCL categories. Decision-level DPO adds $3.1$, $2.2$, and $3.4$ points on the three aggregates. The changes are category-dependent: $\tau^2$ Retail improves by $6.1$ points and BFCL Missing-Parameter by $8.0$, while Missing-Function decreases by $1.5$. Thus, SFT provides the primary capability gains, while the aggregate DPO gains from $1{,}150$ pairs support reusing the same restorable states and executable verifiers for post-SFT decision refinement.

\textbf{Comparison with reference models.}
Davinci-Tool-30B exceeds Qwen3-Coder-480B-A35B-Instruct on all three aggregates despite using $1/16$ as many total parameters. On BFCL v4, its $62.8$ score exceeds Qwen3-Coder-480B, Kimi-K2-Instruct, GPT-5.2, and Claude-Sonnet-4.5, while remaining below MiniMax-M2.1 and GLM-4.7. On MCP-Atlas, its $56.2$ exceeds Qwen3-Coder-480B and GLM-4.7 but remains below the other evaluated reference models.

\textbf{Backbone transfer and data comparison.}
Davinci-ToolData improves the tested smaller models: Qwen3-8B rises from $32.5$ to $74.6$ on $\tau^2$-bench and from $15.4$ to $33.9$ on BFCL v4, while Qwen3-14B rises from $33.4$ to $76.8$ and from $21.1$ to $41.4$, respectively. Both improve in every reported $\tau^2$ and BFCL category, indicating that the supervision is not specific to the primary 30B model. Under the same Qwen3-Coder-30B backbone and SFT hyperparameters, Davinci-ToolData exceeds Toucan-1.5M by $7.4$, $18.7$, and $30.3$ points on $\tau^2$-bench, BFCL v4, and MCP-Atlas, respectively, despite using a substantially smaller training corpus. This matched comparison shows that corpus size alone does not explain the gains.

\textbf{Execution-level analysis.}
Appendix~\ref{app:execution-analysis} compares paired base/SFT trajectories on recovered tasks and reveals benchmark-specific behavioral changes. On $\tau^2$-bench, recovered tasks show substantially fewer explicit tool errors; on BFCL v4, trajectories cover more of the reference call chain and exhibit stronger output-to-argument dependencies; on MCP-Atlas, empty retrievals and malformed terminal outputs disappear despite increased tool interaction. These patterns show that the benchmark gains are associated with improved execution reliability, multi-step completeness, and result utilization rather than uniformly shorter or simpler trajectories.

\subsection{Source-Composition Analysis}
\label{sec:ablations}

\begin{wraptable}[12]{r}{0.57\textwidth}
\vspace{-1.0em}
\centering
\scriptsize
\caption{SFT data-source mixtures. Bold marks the strongest evaluated mixture on each benchmark.}
\label{tab:data-ablation}
\vspace{0.2em}
\setlength{\tabcolsep}{2.6pt}
\renewcommand{\arraystretch}{0.96}
\begin{tabular}{@{}cccccc@{}}
\toprule
\multicolumn{3}{c}{\textbf{SFT data}}
& \multicolumn{3}{c}{\textbf{Benchmarks}} \\
\cmidrule(lr){1-3}
\cmidrule(lr){4-6}
\textbf{EGTS}
& \textbf{EES-Single}
& \textbf{EES-Cross}
& $\boldsymbol{\tau^2}$\textbf{-bench}
& \textbf{BFCL v4}
& \textbf{MCP-Atlas} \\
\midrule
$\checkmark$ & &
& 78.9 & 57.4 & 28.1 \\
& $\checkmark$ &
& 82.0 & 54.5 & 29.2 \\
& & $\checkmark$
& 83.4 & 46.6 & 47.2 \\
& $\checkmark$ & $\checkmark$
& 82.9 & 55.6 & 50.6 \\
$\checkmark$ & $\checkmark$ &
& 79.4 & 57.6 & 31.5 \\
$\checkmark$ & $\checkmark$ & $\checkmark$
& \textbf{85.1} & \textbf{60.6} & \textbf{52.8} \\
\bottomrule
\end{tabular}
\vspace{-0.6em}
\end{wraptable}

Table~\ref{tab:data-ablation} compares six SFT mixtures that differ in both source composition and corpus size, so we use them to characterize source complementarity rather than size-controlled marginal effects. EGTS gives the strongest single-source BFCL v4 result ($57.4$), while EES-Cross is strongest on $\tau^2$-bench ($83.4$) and MCP-Atlas ($47.2$). These distinct profiles support complementary roles for authored-system replay and synthesized-domain supervision.

Source combinations are not uniformly additive: some pairwise mixtures underperform a constituent single source. Nevertheless, adding EES-Cross to EGTS+EES-Single improves $\tau^2$-bench, BFCL v4, and MCP-Atlas by $5.7$, $3.0$, and $21.3$ points, respectively, while adding EGTS to EES-Single+EES-Cross yields gains of $2.2$, $5.0$, and $2.2$ points. The full EGTS+EES-Single+EES-Cross mixture is strongest among all six configurations, exceeding the best single-source result by $1.7$, $3.2$, and $5.6$ points on the three benchmarks, respectively.

Appendix~\ref{app:trajectory-alignment} shows benchmark-dependent source alignment: EGTS is most aligned with BFCL v4 and MCP-Atlas, while EES-Single is most aligned with $\tau^2$-bench, supporting non-redundant source roles alongside Table~\ref{tab:data-ablation}. Appendix~\ref{app:cross-structure-analysis} further shows that, after matching tool-call count and interaction length, EES-Cross retains domain transitions, cross-domain entity reuse, and read--write dependencies, indicating coordinated cross-domain structure beyond simply longer trajectories.

%% file: sections/conclusion.tex
\section{Conclusion}

We presented Davinci-Tool and Davinci-ToolData, an execution-grounded framework for scaling verified supervision for agentic tool use. EGTS converts explored outcomes in independently authored systems into verified task-conditioned demonstrations, while EES constructs audited executable domains for single- and cross-domain supervision. Across three held-out benchmarks, verified-demonstration SFT provides the primary gains and transfers to the tested smaller models. Restorable EES states and executable verifiers are further reused for outcome-grounded decision refinement through DPO. Together, these results show how reusable execution and verification infrastructure can support tool-use training from demonstration construction to decision refinement.

%% file: sections/appendix.tex

\renewcommand{\topfraction}{0.95}
\renewcommand{\bottomfraction}{0.90}
\renewcommand{\textfraction}{0.06}
\renewcommand{\floatpagefraction}{0.68}
\setcounter{topnumber}{3}
\setcounter{bottomnumber}{2}
\setcounter{totalnumber}{5}
\setlength{\textfloatsep}{12pt}
\setlength{\floatsep}{10pt}
\setlength{\intextsep}{10pt}

\section{Shared Execution and Verification Interface}
\label{app:shared-evidence}

This appendix specifies the shared execution, restoration, and verification interface used by all trajectory-construction routes.

\subsection{Environment Interface and Evidence Records}

A resettable tool environment is represented as $W=(\mathcal{S},\mathcal{T},\rho,P,\mathcal{E})$, where $\mathcal{S}$ is the state space, $\mathcal{T}$ the tool set, $\rho$ the restoration operation, $P$ the tool-conditioned transition function, and $\mathcal{E}$ the observable evidence space. For a task $q$, a trajectory $\tau=(a_1,o_1,\ldots,a_H,o_H)$ maps an initial state $s_0$ to a terminal state $s_H$. Its evidence record
\[
R=(q,s_0,\tau,s_H,\xi,\nu)
\]
retains outcome evidence $\xi\in\mathcal{E}$ together with verification and provenance metadata $\nu$.

Evidence capture follows the environment type. Mutation-oriented environments retain state snapshots, retrieval-oriented environments retain grounded answer evidence, and synthesized EES scenarios evaluate executable task predicates. The shared interface supports repeated execution, state restoration, and outcome-evidence capture as required by each construction route. EES scenarios additionally retain scenario-specific executable verifiers that are reused for preference construction in Appendix~\ref{app:dpo-details}.

\subsection{Environment Wrapping and Evidence Capture}

Executable environments are adapted to a common interaction interface exposing callable tools, a reset procedure, outcome-evidence capture, and collection metadata. Depending on the underlying system, restoration may use a container lifecycle, filesystem image, database snapshot, service endpoint, or another source-specific mechanism.

Before each collection episode, the harness records the initial state $s_0$. During execution, it records ordered actions, observations, and execution status:
\[
\tau=
\bigl[
(a_1,o_1,z_1),
\ldots,
(a_H,o_H,z_H)
\bigr],
\]
where $z_t$ indicates whether the corresponding tool invocation completed successfully. After execution, the harness records the resulting state $s_H$ and observable outcome evidence $\xi$.

For state-modifying environments, source-specific snapshot adapters map raw states to canonical representations. Volatile fields such as timestamps, transient identifiers, and execution metadata are removed when they do not affect task semantics, enabling consistent comparison across repeated executions. Retrieval-oriented environments instead retain grounded information returned by successful read operations as answer evidence. EGTS uses these records for task synthesis and replay verification, while EES combines similar state and evidence records with executable task predicates.

\section{Trajectory Construction Details}
\label{app:trajectory-details}

This section provides implementation details for EGTS, EES-Single, and EES-Cross under the shared interface.

\subsection{Execution-Grounded Task Synthesis}
\label{app:egts-details}

\subsubsection{Progressive and Experience-Guided Exploration}

\textbf{Exploration objectives.}
Exploration is organized around local executable objectives derived from the currently exposed tools and observable environment structure. These objectives guide the collector toward feasible interaction patterns, while coverage statistics prioritize under-explored tools and behaviors to broaden the outcomes available for task synthesis.

\textbf{Progressive action-space expansion.}
EGTS progressively expands the action space available to the explorer. Exploration begins with orientation tools that reveal environment structure, proceeds to task-bearing anchor tools, and then expands to the broader tool inventory. As coverage accumulates, under-used tools receive higher sampling priority to encourage new interaction patterns.

Two additional sources of guidance support exploration. Skill guides provide reusable procedural knowledge for common workflows, while an experience store retrieves relevant execution traces from previous episodes. Together, they transfer useful execution knowledge across exploration objectives and support more involved interactions~\citep{yang2025exif,ramrakhya2025autoplay}.

Collection behavior adapts to interaction difficulty. Simpler stages use strict execution, where an unsuccessful call terminates the episode, while harder stages permit bounded retry and rollback. If execution ends with an unsuccessful tail, the collector retains the longest executable prefix ending at a successful call and removes unmatched or failure-only messages.

\subsubsection{Task Candidate Construction}

For mutation-oriented exploratory trajectories, task synthesis conditions on the retained calls, affected entities, and observed transition from $s_0$ to $s_H$. For retrieval-oriented trajectories, it conditions on successful reads and their grounded answer evidence. The synthesizer generates outcome-oriented tasks supported by the recorded execution~\citep{wang2026trajectory2task,ouajdi2026executionfirst}.

Multiple task candidates may be generated from one exploratory interaction. Candidate selection considers specificity, naturalness, and consistency with the recorded evidence, retaining requests whose target outcomes can be traced to the corresponding calls and observations.

\subsubsection{Execution-Derived Replay Guidance}

The replay harness associates the observed outcome with the calls that contributed to it and extracts task-relevant execution information from the exploratory episode. This yields a replay guide
\[
g(\tau_{\mathrm{exp}},\xi_{\mathrm{exp}})
\]
used during replay:
\[
s_0
\xrightarrow[\pi_{\mathrm{rep}}]
{q,\,g(\tau_{\mathrm{exp}},\xi_{\mathrm{exp}})}
\tau_{\mathrm{rep}},s'_H.
\]

The guide transfers grounded procedural information from exploration to task-conditioned replay~\citep{xu2025agenttrek}. Unlike an exploratory trace, the replay trajectory is generated after the user request has been fixed and is therefore organized around fulfilling that request.

Replay is evaluated as part of demonstration construction. Its acceptance records whether the synthesized request recovers the target outcome under the replay harness.

The EGTS SFT partition contains $13{,}824$ accepted task-conditioned replay demonstrations. Each serialized example contains \texttt{messages} and \texttt{tools} and uses the standard task-solving system prompt. Replay guidance is privileged collection context and is not serialized as a separate training field; the collection prompt is replaced during SFT serialization. The resulting training example retains the task-conditioned interaction rather than the exploration-oriented collection context.

\subsubsection{Replay Verification and Trajectory Audit}

For mutation tasks, replay succeeds when the canonical terminal state matches the outcome recorded during exploration:
\[
\phi_W(s'_H)=\phi_W(s_H).
\]
Retrieval tasks are evaluated against the grounded answer evidence collected during exploration.

After outcome verification, additional checks determine whether the replayed trajectory is suitable for training. Structural validation enforces valid message--tool-call pairing, executable tool invocation, final-response presence, and failure constraints. A hard judge checks consistency among the task, tool names, arguments, observations, action ordering, and final evidence, while a soft judge evaluates trajectory coherence and response quality.

Accepted samples retain the task, reset provenance, execution snapshots, replay trace, task-candidate records, harness metadata, and verification results, linking demonstrations to acceptance decisions.

\subsection{Executable Environment Synthesis}
\label{app:programmatic-details}

This section expands the EES procedures introduced in Sections~\ref{sec:pes-single} and~\ref{sec:pes-cross}, covering single-domain synthesis and auditing, state-grounded scenario construction, executable domain composition, and cross-domain trajectory verification.

\subsubsection{Single-Domain Synthesis and Audit}

\textbf{Staged contract synthesis.}
For each target domain, EES first mines representative backend concepts, entity structures, operational constraints, and interaction patterns. After deduplication, staged synthesis converts this information into typed schemas, executable scaffolds, operation inventories, and concrete tool implementations, producing a runnable domain with explicit state, transition, and rule structure~\citep{song2026envscaler}.

The resulting domain contract specifies entity types, state structure, exposed operations, transition behavior, and domain rules. This contract provides shared semantics for tool execution, scenario construction, and verification.

\textbf{Executor--auditor protocol.}
Each synthesized domain is exercised before scenario construction. The executor invokes its operations over initialized states, covering both state-reading and state-mutating behavior. The auditor checks tool arguments, returned values, exceptions, resulting state transitions, and rule compliance against the domain contract.

Detected inconsistencies trigger revisions to the corresponding schema, implementation, or domain specification before another audit round. Audited domains are used for scenario construction.

\subsubsection{State-Grounded Task and Verifier Construction}

\textbf{Scenario initialization and task construction.}
For each audited domain, EES instantiates an initial state $s_0$ consistent with its entity schemas and domain rules. Task generation conditions on this state, the exposed operations, and applicable constraints to construct achievable interaction requirements. These requirements are decomposed into atomic conditions represented by executable predicates.

Each resulting scenario has the form
\[
(s_0,q,\mathcal{V}),
\]
where $q$ is the state-grounded user task and
\[
\mathcal{V}=\{v_1,\ldots,v_J\}
\]
contains executable predicates encoding its atomic requirements.

For state-modifying tasks, state-based target predicates are first evaluated on the initial state $s_0$. A target already satisfied in $s_0$ is regenerated or revised so that completing the task requires an interaction-induced state change. Candidate trajectories are then collected by executing $q$ from $s_0$, producing an execution trace $\tau$ and terminal state $s_H$. The task requirements are evaluated on the resulting execution and state:
\[
\operatorname{Verify}(q,\tau,s_H)
=
\prod_{j=1}^{J}
\mathbb{I}
\left[
v_j(\tau,s_H)=1
\right].
\]

The initial state, task requirements, execution trace, terminal state, predicate outcomes, and verification record form the executable evidence retained with each verified EES-Single trajectory.

\subsubsection{Cross-Domain Composition}

\textbf{Affinity estimation and subgraph sampling.}
For each audited domain, embedding retrieval proposes semantically related candidates. An affinity rubric then evaluates their functional compatibility and possible relationships between their entity and state spaces. These relations define an affinity graph; bounded random walks sample connected domain subsets for composition~\citep{li2026toolathlon}.

\textbf{Namespace injection and executable merging.}
For each selected domain $d_i$, namespace injection $\pi_i$ transforms tool names, entity types, identifiers, and state attributes before composition. The merged environment state is
\[
s^{\mathrm{cross}}
=
\bigl(
\pi_1(s_1),
\ldots,
\pi_m(s_m)
\bigr),
\]
and the exposed tool set is formed from the corresponding namespaced interfaces.

Static validation checks initialization, state structure, tool-schema well-formedness, and namespace consistency before the merged runtime proceeds to shared world-context construction.

\textbf{Shared world-context construction and audit.}
EES-Cross constructs a shared world context over the merged domains. The context specifies user roles, shared entities, linking fields, cross-domain policies, and coherence constraints that establish relationships among the participating domain states.

Schema and consistency checks validate referenced entities, attributes, and interfaces against the merged runtime. A consistency judge further evaluates the generated roles, policies, linking relations, and shared entities. Accepted world contexts are then used for cross-domain task construction.

\subsubsection{Cross-Domain Task Construction and Verification}

Merged environments are initialized with records whose identifiers and linking fields establish dependencies across participating domains. Cross-domain tasks are conditioned on the shared world context, user role, initialized state, and applicable policies, with required operations and success criteria spanning multiple systems.

Candidate trajectories first pass runtime and tool-execution checks. Task-level verification then evaluates the observed calls, state changes, terminal outcome, and applicable policy constraints against the scenario requirements. Trajectories satisfying these requirements are retained.

Accepted trajectories retain the initial and terminal states, shared world context, cross-domain execution trace, outcome evidence, verification records, transferred-entity types, and dependency structure together with their construction and verification provenance.

\section{Environment and Data Analysis}
\label{app:environment-data-analysis}

\subsection{Environment Sources and Coverage}
\label{app:worlds}

This section characterizes the authored environment families used by EGTS and the domain taxonomy and composition patterns used by EES. Trajectory-level statistics for the resulting SFT corpus are reported in Section~\ref{sec:corpus-statistics} and Appendix~\ref{app:corpus-statistics}.

\subsubsection{EGTS Environment Sources}

EGTS operates over independently authored environments that support multi-step tool interaction, repeatable initialization, observable state or answer evidence, and stable execution. Table~\ref{tab:world-categories} summarizes the represented environment families and their interaction scope.

\begin{table}[!htbp]
\centering
\small
\caption{Environment families represented in the EGTS library. Each source is adapted to the common execution and evidence-capture interface.}
\label{tab:world-categories}
\resizebox{\textwidth}{!}{%
\begin{tabular}{
>{\raggedright\arraybackslash}p{0.22\textwidth}
>{\raggedright\arraybackslash}p{0.42\textwidth}
>{\raggedright\arraybackslash}p{0.28\textwidth}}
\toprule
\textbf{Environment family}
& \textbf{Representative interactions}
& \textbf{Representative adapted sources} \\
\midrule

Personal and business services
& Contacts, messages, reminders, calendars, banking, travel, health, and other structured service operations
& ToolSandbox~\citep{lu2024toolsandbox}, API-Bank~\citep{li2023apibank} \\

API discovery and orchestration
& API search, schema inspection, argument grounding, selection, and multi-API composition
& StableToolBench~\citep{guo2024stabletoolbench} \\

Developer and data workspaces
& Filesystem operations, repositories, relational data, and browser-backed developer workflows
& MCPMark~\citep{wu2025mcpmark} \\

Terminal and professional production
& Shell and text processing, system operations, reports, office files, PDFs, and data deliverables
& Terminal-Bench~\citep{merrill2026terminalbench} \\

Web interaction and commerce
& Page navigation, form interaction, product search, constraint matching, and shopping workflows
& MiniWoB~\citep{shi2017worldofbits}, WebShop~\citep{yao2022webshop} \\

Scientific simulation
& Procedural experiments, hypothesis testing, embodied discovery, and stateful scientific interaction
& ScienceWorld~\citep{wang2022scienceworld}, DiscoveryWorld~\citep{jansen2024discoveryworld} \\

Integrated multi-tool workflows
& Workspace, browser, research, document, communication, commerce, and infrastructure tools within a shared environment
& Toolathlon~\citep{li2026toolathlon} \\

Application ecosystems
& Stateful workflows spanning multiple simulated applications and user records
& AppWorld~\citep{trivedi2024appworld} \\

\bottomrule
\end{tabular}%
}
\end{table}

The collector synthesizes new tasks from explored outcomes while retaining environment identifiers, source versions, reset provenance, and task-level metadata. Evaluation queries and states from BFCL v4, $\tau^2$-bench, and MCP-Atlas are excluded from trajectory construction.

\textbf{Scope of the source inventory.}
Table~\ref{tab:world-categories} describes the authored-system library, whereas Appendix~\ref{app:corpus-statistics} characterizes the $13{,}824$ accepted EGTS demonstrations used for SFT. Environment-family counts and trajectory counts therefore describe different levels of the construction pipeline.

\subsubsection{EES Domain Taxonomy and Usage}
\label{app:ees-coverage}

The framework design and data in EES-Single are adapted from prior work~\citep{song2026envscaler}. The EES catalog contains 191 executable domain definitions organized into 19 application categories, ranging from healthcare and software infrastructure to commerce, education, and document management. A domain is a stateful executable component; a category groups domains by application area. The retained EES-Cross inventory contains 246 compositions using 188 of the 191 domains, with 706 constituent-domain slots in total.

Table~\ref{tab:ees-domain-categories} reports both the number of domain definitions in each category and their frequency in the retained cross-domain compositions. A domain slot is one occurrence of a constituent domain within one composed environment.

\begin{table}[!htbp]
\centering
\small
\caption{EES domain categories and composition usage. $D_c$ is the number of domain definitions in category $c$, and $S_c$ counts their occurrences across the 246 retained EES-Cross compositions. Rows follow decreasing $S_c$. Category codes are used in Figure~\ref{fig:ees-composition-patterns}.}
\label{tab:ees-domain-categories}
\begin{tabular}{@{}lp{0.62\textwidth}rr@{}}
\toprule
\textbf{Code}
& \textbf{Application category}
& $\boldsymbol{D_c}$
& $\boldsymbol{S_c}$ \\
\midrule
HC & Healthcare \& Clinical & 30 & 115 \\
IT & IT Infrastructure \& Software Engineering & 20 & 80 \\
SOC & Social \& Communication & 14 & 56 \\
ECOM & E-commerce, Retail \& Marketplace & 11 & 47 \\
ENT & Enterprise, HR \& Business Ops & 12 & 47 \\
PHW & Personal Health \& Wellness & 13 & 47 \\
PROD & Productivity \& Personal Info Mgmt & 8 & 45 \\
FIN & Finance, Banking \& Billing & 13 & 40 \\
MEDIA & Media \& Entertainment & 10 & 39 \\
TRAVEL & Travel, Hospitality \& Local Services & 8 & 30 \\
SPORT & Sports \& Fitness & 11 & 27 \\
DATA & Data, Analytics \& ML & 6 & 24 \\
IOT & IoT, Industrial \& Environmental Monitoring & 8 & 22 \\
MKT & Marketing \& Advertising & 4 & 20 \\
DOC & Documents \& Knowledge Mgmt & 4 & 16 \\
INS & Insurance \& Claims & 5 & 16 \\
EDU & Education & 4 & 13 \\
GOV & Government \& Public Sector & 5 & 13 \\
MOB & Transportation, Mobility \& Geo & 5 & 9 \\
\midrule
& Total & 191 & 706 \\
\bottomrule
\end{tabular}
\end{table}

\textbf{Category coverage.}
Healthcare and Clinical accounts for 115 of the 706 constituent-domain slots (16.3\%), followed by IT Infrastructure and Software Engineering with 80 slots (11.3\%) and Social and Communication with 56 (7.9\%). The remaining slots are distributed across all 16 other application categories, showing that the retained compositions span the full EES taxonomy rather than concentrating on a small number of domains.

\subsubsection{EES Composition Patterns and Category Affinities}
\label{app:ees-composition-patterns}

The retained EES-Cross inventory contains 77 two-domain, 124 three-domain, and 45 four-domain compositions, totaling 246 compositions and 706 constituent-domain slots. The mean composition size is 2.87 domains. The category-count distribution also totals 246 compositions: 129 contain domains from one application category, while 117 span multiple categories. Domain count and category count capture different aspects of composition because distinct executable systems can belong to the same application category while retaining separate states and tool interfaces. Figure~\ref{fig:ees-composition-patterns} in Section~\ref{sec:corpus-statistics} visualizes these distributions and the most frequent cross-category affinities.

Healthcare and Clinical with Personal Health and Wellness is the most frequent category pair, appearing in 17 compositions, followed by Data, Analytics and ML with IT Infrastructure and Software Engineering in 15. Other recurring combinations include E-commerce with Travel and Enterprise operations with Productivity, each appearing in six compositions. These patterns illustrate both within-category system composition and workflows connecting distinct application areas.

\subsubsection{Structural Analysis of Cross-Domain Supervision}
\label{app:cross-structure-analysis}

Beyond composition statistics, we further analyze whether EES-Cross
introduces execution structures that differ from single-domain trajectories.
The goal is to distinguish cross-domain supervision from simply longer
interaction sequences. We compare EES-Single and EES-Cross trajectories
under length-controlled matching and measure structural properties related
to cross-system coordination.

For each trajectory, we represent the executed tool sequence as an ordered
chain of tool invocations. We identify domain transitions when two
consecutive tool calls originate from different executable domains. Given a
trajectory $\tau=(a_1,\ldots,a_H)$ with domain assignment
$d(a_t)$ for each tool call, the domain-switch ratio is defined as

\[
R_{\mathrm{switch}}(\tau)
=
\frac{1}{H-1}
\sum_{t=1}^{H-1}
\mathbb{I}
[d(a_t)\neq d(a_{t+1})].
\]

To measure information transfer across domains, we additionally identify
entities produced by one domain and consumed by another domain. For an
entity $e$, let $p(e)$ and $c(e)$ denote the producing and consuming tool
calls, respectively. We count cross-domain entity reuse when
$d(p(e))\neq d(c(e))$. We further identify cross-domain read--write
patterns where a value retrieved from one domain contributes to a later
state-modifying operation in another domain.

Because EES-Cross trajectories may naturally contain more involved
interactions, we perform a length-controlled comparison. For each
EES-Cross trajectory, we select EES-Single trajectories with matched tool-call
counts and similar serialized interaction lengths, and compute the structural
statistics on the matched subsets. Table~\ref{tab:cross-structure-analysis}
reports the resulting comparison.

\begin{table}[htbp]
\centering
\small
\caption{
Structural comparison between EES-Single and EES-Cross trajectories under
length-controlled matching. Cross-domain metrics measure execution
coordination beyond interaction length.
}
\label{tab:cross-structure-analysis}
\begin{tabular}{@{}lcc@{}}
\toprule
\textbf{Metric}
& \textbf{EES-Single}
& \textbf{EES-Cross}
\\
\midrule

Domain transitions (\%)
& 0.00
& 14.50
\\

Cross-domain entity reuse (\%)
& 0.00
& 5.23
\\

Trajectories with cross-domain transfer (\%)
& 0.00
& 6.39
\\

Cross-domain read$\rightarrow$write (\%)
& 0.00
& 0.98
\\

Tool calls (matched)
& 100.0
& 100.0
\\

Interaction length (relative)
& 100.0
& 99.1
\\

\bottomrule
\end{tabular}
\end{table}

The comparison shows that EES-Cross introduces execution patterns absent
from EES-Single, including domain transitions, cross-domain entity transfer,
and read--write dependencies across systems. These differences remain after
controlling for tool-call count and interaction length, indicating that the
additional supervision arises from coordinated execution structures rather
than simply longer trajectories. The analysis does not imply that every
dimension of structural depth increases; instead, EES-Cross changes the
dependency topology of tool-use workflows by introducing interactions among
otherwise independent executable domains.

\subsubsection{Recorded Composition: Inventory, Promotions, and Orders}
\label{app:cross-composition-example}

The composition manifest identifies \texttt{cross\_env\_0020} as a merge of \texttt{env\_9\_sft} (product inventory), \texttt{env\_76\_sft} (product-and-promotion management), and \texttt{env\_172\_rl} (order management). Table~\ref{tab:cross-composition-example} reports selected operations and rules from their domain definitions.

\begin{table}[htbp]
\centering
\small
\caption{Selected interfaces in the recorded composition \texttt{cross\_env\_0020}. Names are shown before namespace injection.}
\label{tab:cross-composition-example}
\begin{tabular}{@{}p{0.16\textwidth}p{0.42\textwidth}p{0.36\textwidth}@{}}
\toprule
\textbf{Domain}
& \textbf{Selected operations}
& \textbf{Declared requirements} \\
\midrule

Inventory
& \texttt{get\_product\_by\_id}; \texttt{update\_product\_stock}
& Non-negative prices; predefined product categories. \\

Promotions
& \texttt{get\_product\_by\_sku}; \texttt{get\_applicable\_promotions}; \texttt{evaluate\_customer\_eligibility}
& Active time window, product--promotion link, and customer eligibility determine applicability. \\

Orders
& \texttt{get\_product\_by\_id}; \texttt{add\_product\_to\_cart}
& Stock must cover the added quantity; each cart belongs to one user; repeated products use an aggregated quantity. \\
\bottomrule
\end{tabular}
\end{table}

The example illustrates two requirements handled separately by EES-Cross. The duplicated \texttt{get\_product\_by\_id} interface motivates namespace separation, while the use of product IDs and SKUs across systems requires semantic links between their underlying records. The shared-context and scenario-construction stages establish these relations before cross-domain task generation.

\subsubsection{Serialized Cross-Domain Demonstration}
\label{app:cross-trajectory-example}

One EES-Cross training example asks the agent to consult the ``Cryo-EM Vendor Call'' note, verify ownership of an existing reminder, update its message and time, and report the user's notification preference. The request names the target note and reminder, while execution retrieves their contents and resolves the corresponding user identity before performing the updates.

\begin{table}[htbp]
\centering
\small
\caption{Model-visible tool evidence from one serialized two-domain EES-Cross demonstration. Construction-level states and verification records are retained separately from the conversation.}
\label{tab:cross-trajectory-example}
\begin{tabular}{@{}cp{0.44\textwidth}p{0.44\textwidth}@{}}
\toprule
\textbf{Step}
& \textbf{Operation}
& \textbf{Relevant observation or update} \\
\midrule
1
& \texttt{mobile\_get\_user\_by\_name}
& Resolves Maya Chen to \texttt{usr-001}; notification preference is push. \\

2
& \texttt{personal\_get\_note\_by\_id}
& Note content: compare service quotes and ask BioScope about January slots. \\

3
& \texttt{mobile\_get\_reminder\_by\_id}
& \texttt{rem-001} belongs to \texttt{usr-001}, with the existing message and time. \\

4
& \texttt{mobile\_update\_reminder\_message}
& Sets ``Call BioScope about January cryo-EM slots''; tool reports success. \\

5
& \texttt{mobile\_update\_reminder\_time}
& Sets 2025-01-17 at 10:00, UTC$-$05:00; tool reports success. \\

6
& \texttt{mobile\_get\_user\_}\allowbreak\texttt{notification\_preference}
& Queries \texttt{usr-001}; result is push. \\
\bottomrule
\end{tabular}
\end{table}

The example makes the cross-system dependency concrete. Information retrieved from the personal-note interface contributes to a reminder update, while identity resolution provides the user identifier needed to verify reminder ownership and retrieve notification settings.

\subsection{Detailed Corpus Statistics}
\label{app:corpus-statistics}

Davinci-ToolData contains $30{,}395$ SFT trajectories: $13{,}824$ from EGTS, $8{,}026$ from EES-Single, and $8{,}545$ from EES-Cross. Each stored conversation is counted as one task trajectory. Distinct requests are identified from the complete sequence of user messages after Unicode NFKC and whitespace normalization; executable scenarios additionally depend on their initial states. The three source partitions concatenate to the full corpus with no duplicate canonical JSON records. As sets of declared tool names, inventories characterize serialized configurations rather than trajectory counts.

\begin{table}[htbp]
\centering
\small
\caption{Coverage of the SFT corpus. ``Requests'' counts distinct normalized user-message sequences; ``Inventories'' counts distinct declared-tool configurations. Declared and invoked tools are counted by their serialized names.}
\label{tab:corpus-coverage-detailed}
\resizebox{\textwidth}{!}{%
\begin{tabular}{lrrrrr}
\toprule
\textbf{Source}
& \textbf{Trajectories}
& \textbf{Requests}
& \textbf{Inventories}
& \textbf{Declared tools}
& \textbf{Invoked tools} \\
\midrule
EGTS & $13{,}824$ & 13,283 & 117 & 387 & 350 \\
EES-Single & $8{,}026$ & 7,550 & 234 & 4,042 & 1,610 \\
EES-Cross & $8{,}545$ & 8,319 & 276 & 4,316 & 1,683 \\
\midrule
Full SFT corpus & $30{,}395$ & 28,985 & 436 & $4{,}672$ & $2{,}260$ \\
\bottomrule
\end{tabular}%
}
\end{table}

The $4{,}672$ declared tool names describe the union of serialized tool schemas exposed across the corpus, whereas $2{,}260$ distinct serialized tool names are actually invoked in the retained trajectories. The latter is the tool-coverage statistic reported in the main text.

\textbf{Length and interaction definitions.}
Token counts use the SFT checkpoint's tokenizer and chat template, including the system prompt, tool definitions, and complete conversation. JSON-encoded arguments and schemas are parsed before template rendering. A tool call is one explicit assistant tool invocation, so parallel calls count separately. An assistant turn is one assistant message, including messages without tool calls.

\begin{table}[htbp]
\centering
\small
\caption{Length and interaction statistics of the SFT corpus. Token lengths describe complete serialized examples before training-time length handling.}
\label{tab:corpus-lengths}
\resizebox{\textwidth}{!}{%
\begin{tabular}{lrrrrrrrrr}
\toprule
& \multicolumn{3}{c}{\textbf{Tokens}}
& \multicolumn{3}{c}{\textbf{Tool calls}}
& \multicolumn{3}{c}{\textbf{Assistant turns}} \\
\cmidrule(lr){2-4}
\cmidrule(lr){5-7}
\cmidrule(lr){8-10}
\textbf{Source}
& \textbf{Mean}
& \textbf{Min.}
& \textbf{Max.}
& \textbf{Mean}
& \textbf{Min.}
& \textbf{Max.}
& \textbf{Mean}
& \textbf{Min.}
& \textbf{Max.} \\
\midrule

EGTS
& 10,063.9 & 377 & 768,427
& 6.04 & 0 & 60
& 9.80 & 1 & 45 \\

EES-Single
& 10,372.7 & 2,030 & 369,175
& 5.87 & 0 & 76
& 10.42 & 1 & 33 \\

EES-Cross
& 20,444.7 & 1,746 & 1,259,991
& 7.01 & 0 & 99
& 7.57 & 1 & 34 \\
\midrule

Full SFT corpus
& 13,063.8 & 377 & 1,259,991
& 6.26 & 0 & 99
& 9.34 & 1 & 45 \\
\bottomrule
\end{tabular}%
}
\end{table}

\textbf{Length distribution.}
The full SFT corpus contains $190{,}410$ explicit tool calls and averages 4.05 distinct invoked tools per trajectory. Its median serialized length is 9,284 tokens and its 95th percentile is 20,237 tokens. A total of 957 trajectories ($3.15\%$) exceed $32{,}768$ serialized tokens before training-time length handling. EES-Cross has a median length of 10,709 tokens and a mean of 20,444.7, reflecting a longer upper tail than the other two sources. During SFT, examples exceeding the $32{,}768$-token maximum sequence length are truncated by the default training pipeline. The statistics in Table~\ref{tab:corpus-lengths} describe the complete serialized examples before this truncation. The SFT training configuration is reported in Appendix~\ref{app:training-config}.

\subsection{Representation Alignment with Evaluation Trajectories}
\label{app:trajectory-alignment}

To examine how the three construction routes relate to the execution patterns required by held-out evaluations, we compare the complete verified-trajectory files used for this analysis with benchmark-side execution traces. Each synthetic and benchmark trajectory is normalized into three views: the task intent, the available tool capabilities, and the executed action chain. Instance-specific identifiers, entity names, paths, and values are replaced with typed placeholders. We use the executed-action view for the primary comparison and retain the intent and tool views as auxiliary checks. On the benchmark side, we use all official BFCL v4 multi-turn reference chains and all fully successful trajectories in the supplied $\tau^2$-bench and MCP-Atlas evaluation records.

Let $D_s$ denote the trajectories from source $s$ and let $e(x)$ be the Qwen3-Embedding-8B representation of normalized trajectory $x$. For benchmark trajectory $b$, we measure source alignment by the mean cosine similarity of its $K$ nearest source trajectories:
\[
A_s(b)
=
\frac{1}{K}
\sum_{x\in\operatorname{NN}_K(b;D_s)}
\frac{e(b)^\top e(x)}{\lVert e(b)\rVert_2\lVert e(x)\rVert_2},
\qquad K=5.
\]
All sources use the same normalizer, encoder, and pooling procedure. To prevent a larger source from receiving an artificial nearest-neighbor advantage, we deduplicate exact representations and sample equally sized candidate pools from EGTS, EES-Single, and EES-Cross over ten repeats. Table~\ref{tab:trajectory-alignment} reports the resulting mean Top-5 alignment. This metric characterizes relative proximity in the representation space rather than absolute task coverage or execution correctness.

\begin{table}[htbp]
\centering
\small
\caption{Representation alignment between verified synthesis trajectories and benchmark reference or successful execution trajectories. Values are mean Top-5 cosine similarities under equal candidate budgets; bold marks the strongest source for each benchmark.}
\label{tab:trajectory-alignment}
\begin{tabular}{@{}lccc@{}}
\toprule
\textbf{Source}
& \textbf{BFCL v4}
& $\boldsymbol{\tau^2}$\textbf{-bench}
& \textbf{MCP-Atlas} \\
\midrule
EGTS       & \textbf{0.591} & 0.665          & \textbf{0.680} \\
EES-Single & 0.547          & \textbf{0.695} & 0.567          \\
EES-Cross  & 0.575          & 0.670          & 0.589          \\
\bottomrule
\end{tabular}
\end{table}

The source profiles are benchmark dependent. EGTS is most closely aligned with BFCL v4 and the supplied MCP-Atlas trajectories, whereas EES-Single is most closely aligned with $\tau^2$-bench. EES-Cross is not the strongest source in aggregate, but provides additional high-ranked neighbors on BFCL v4 and $\tau^2$-bench. An equal-budget leave-one-source-out comparison gives the same pattern: removing EGTS reduces alignment on BFCL v4 and MCP-Atlas, removing EES-Single reduces alignment on $\tau^2$-bench, and removing EES-Cross produces a smaller reduction on BFCL v4. These results indicate that the construction routes contribute different execution-pattern neighborhoods. Together with the source-mixture results in Table~\ref{tab:data-ablation}, they support the interpretation that the full mixture combines non-redundant supervision; downstream training benefits are evaluated by the source-mixture experiments rather than by embedding similarity alone.

\subsection{Execution-Level Analysis of Benchmark Gains}
\label{app:execution-analysis}

We further examine whether the benchmark gains after SFT are accompanied by
changes in actual tool execution. For each benchmark, we align the base-model
and Davinci-Tool-30B-SFT trajectories by task identity and retain the original
benchmark success criterion. Let $M_b$ denote the matched task set for benchmark
$b$, and let $y_i^{\mathrm{base}},y_i^{\mathrm{sft}}\in\{0,1\}$ denote the
outcomes of task $i$ before and after SFT. We summarize task-level transitions
using
\[
R_{u\rightarrow v}(b)
=
\frac{1}{|M_b|}
\sum_{i\in M_b}
\mathbb{I}
\left[
y_i^{\mathrm{base}}=u,\;
y_i^{\mathrm{sft}}=v
\right],
\]
where $u,v\in\{0,1\}$. Table~\ref{tab:execution-transition-summary} reports the
resulting transition counts. Across 1,116 matched tasks, 441 change from failure
to success (F$\rightarrow$S), while 78 change from success to failure
(S$\rightarrow$F).

\begin{table}[htbp]
\centering
\small
\caption{
Paired outcome transitions between the base model and Davinci-Tool-30B-SFT.
}
\label{tab:execution-transition-summary}
\begin{tabular}{@{}lrrrrr@{}}
\toprule
\textbf{Benchmark}
& \textbf{Matched}
& \textbf{F$\rightarrow$S}
& \textbf{S$\rightarrow$F}
& \textbf{S$\rightarrow$S}
& \textbf{F$\rightarrow$F} \\
\midrule
$\tau^2$ Retail  & 114 & 23  & 10 & 68  & 13  \\
$\tau^2$ Telecom & 113 & 74  & 4  & 35  & 0   \\
BFCL v4          & 800 & 304 & 58 & 181 & 257 \\
MCP-Atlas        & 89  & 40  & 6  & 9   & 34  \\
\bottomrule
\end{tabular}
\end{table}

\textbf{Execution statistics.}
We analyze the F$\rightarrow$S trajectories to characterize how recovered tasks
change at the execution level. Because the benchmarks expose different tool
interfaces and success conditions, we use benchmark-specific observable
statistics rather than forcing a single execution-quality metric across all
tasks. For $\tau^2$-bench, we measure the explicit tool-error rate
\[
\operatorname{ErrRate}(\tau)
=
\frac{N_{\mathrm{err}}(\tau)}
     {N_{\mathrm{call}}(\tau)},
\]
where $N_{\mathrm{err}}$ and $N_{\mathrm{call}}$ are the numbers of explicit
tool errors and tool calls in trajectory $\tau$.

For BFCL v4, we additionally measure how completely the executed trajectory
covers the normalized reference-call chain. Let $C_i^{\mathrm{ref}}$ denote the
normalized reference call signatures for task $i$. We define
\[
\operatorname{Cov}(\tau_i)
=
\frac{
N_{\mathrm{match}}(\tau_i,C_i^{\mathrm{ref}})
}{
|C_i^{\mathrm{ref}}|
},
\]
where $N_{\mathrm{match}}$ counts executed calls matched to the normalized
reference signatures. We further construct conservative output-to-argument
dependencies when a value returned by an earlier successful call is reused in
a later call argument, excluding values already present in the user request,
input echoes, and trivial constants. From this graph we report the number of
visible dependency edges and its longest visible chain. BFCL behavior statistics
use the 274 recovered tasks with executable calls available on both sides;
missing execution traces are left unknown rather than treated as zero.

For MCP-Atlas, we use directly observable execution-profile statistics,
including empty retrieval responses, protocol-fragment endings, calls per task,
and explicit tool-error rates. These measurements characterize the executed
trajectory and are not used as alternative benchmark scores.

\begin{figure}[htbp]
\centering
\begin{subfigure}[t]{0.485\linewidth}
    \centering
    \includegraphics[width=\linewidth]{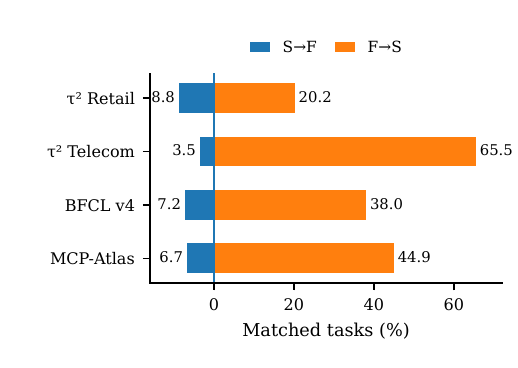}
    \caption{Outcome transitions.}
\end{subfigure}\hfill
\begin{subfigure}[t]{0.485\linewidth}
    \centering
    \includegraphics[width=\linewidth]{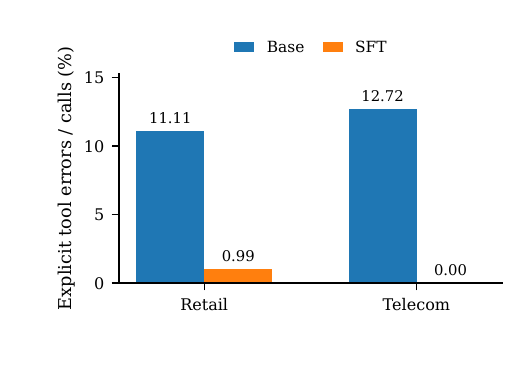}
    \caption{$\tau^2$ reliability.}
\end{subfigure}
\begin{subfigure}[t]{0.485\linewidth}
    \centering
    \includegraphics[width=\linewidth]{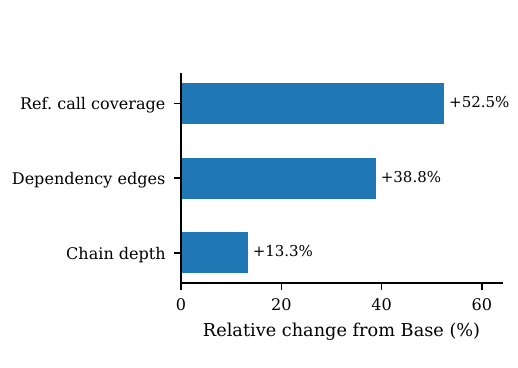}
    \caption{BFCL completeness.}
\end{subfigure}\hfill
\begin{subfigure}[t]{0.485\linewidth}
    \centering
    \includegraphics[width=\linewidth]{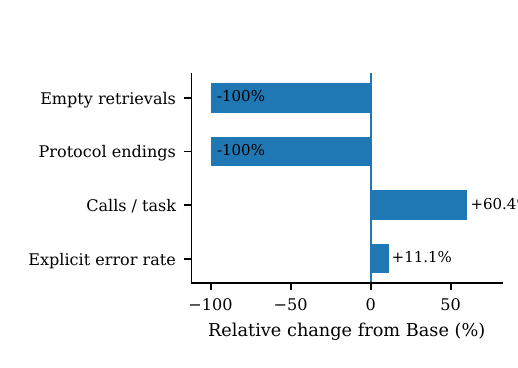}
    \caption{MCP execution profile.}
\end{subfigure}

\caption{
Paired execution changes associated with benchmark recovery after SFT.
(a) shows the fractions of matched tasks changing from failure to success and
from success to failure.
(b) shows that, among recovered $\tau^2$-bench tasks, explicit tool-error rates
decrease from 11.11\% to 0.99\% on Retail and from 12.72\% to 0 on Telecom.
(c) shows that recovered BFCL v4 trajectories become more complete:
normalized reference-call coverage increases from 56.41\% to 86.01\%, visible
output-to-argument dependency edges from 1.21 to 1.68, and visible chain depth
from 1.58 to 1.79.
(d) shows a different pattern on MCP-Atlas: observed empty retrievals decrease
from 37 to 0 and protocol-fragment endings from 7 to 0, while calls per task
increase from 5.55 to 8.90 and the explicit tool-error rate changes from
14.41\% to 16.01\%.
The paired analysis therefore indicates benchmark-specific execution changes
rather than a single mechanism based on shorter trajectories or uniformly
fewer tool errors.
}
\label{fig:execution-recovery-analysis}
\end{figure}

\textbf{Observed behavior changes.}
Figure~\ref{fig:execution-recovery-analysis} shows that the performance gains
are accompanied by concrete but different changes in execution behavior.
On $\tau^2$-bench, recovery is associated with substantially more reliable tool
execution; Retail additionally recovers the expected database state on 18 of
23 F$\rightarrow$S tasks, while the 18 observed Telecom identity-placeholder
calls disappear after SFT. BFCL v4 exhibits a different pattern: recovered
trajectories execute more of the reference chain and expose stronger
output-to-argument dependencies, even though the average number of calls
increases from 7.62 to 9.57. MCP-Atlas likewise does not show a general reduction
in execution cost; instead, recovered trajectories avoid observed dead-end
retrieval loops and malformed terminal outputs while performing more tool
interaction. These results complement the aggregate benchmark improvements by
showing that Davinci-ToolData is associated with observable gains in execution
reliability, multi-step completeness, and result utilization rather than a
uniform reduction in trajectory length or tool use.

This analysis is descriptive rather than a separate causal estimate: each
model contributes one recorded rollout per task, and F$\rightarrow$S follows
the original benchmark scoring rule, which can include partially correct
outcomes.

\section{Same-Prefix Preference Construction Details}
\label{app:dpo-details}

This appendix expands the preference-construction procedure introduced in Section~\ref{sec:preference-construction}. Preference pairs are constructed after SFT from restorable EES-Single and EES-Cross scenarios using the SFT checkpoint for candidate generation and continuation rollout. We report the collection settings used for the primary preference corpus; engineering-only execution settings and exploratory collection variants are omitted.

\subsection{Fork-State Collection}

For each selected EES-Single or EES-Cross scenario, the SFT policy first generates an interaction rollout, referred to as the trunk. Immediately before each assistant decision, the collector records the restorable environment state together with the interaction prefix and visible tool schemas. We retain at most $B=10$ recorded decision points along each trunk as candidate fork states. The trunk is generated greedily with temperature $0$, providing a deterministic reference path for sibling construction.

A recorded fork is
\[
h_t=(q,s_t,m_{<t},\mathcal{T}_t),
\]
where $q$ is the user task, $s_t$ is the executable state immediately before the decision, $m_{<t}$ is the shared model-visible interaction history, and $\mathcal{T}_t$ contains the tools exposed at that point.

\subsection{Sibling Generation and Rollout}

At each selected fork, the assistant turn from the trunk is retained as one candidate and $K-1$ alternative decisions are sampled from the same SFT policy. For the preference data used in the main experiments, we set $K=3$, yielding one greedy trunk decision and two sampled sibling decisions per fork. Stochasticity is introduced only at the fork decision: sibling decisions use temperature $1.0$, while the trunk decision and all subsequent continuations use temperature $0$ (greedy decoding). Branch samples use deterministic branch-specific seeds derived from a base seed of 1000.

Before executing each candidate, the collector restores the same environment state $s_t$ and reuses the same task, interaction prefix, and visible tool set. Each candidate decision $u_t^{(k)}$ is executed from the recorded state $s_t$ under the shared model-visible context $x_t=(q,m_{<t},\mathcal{T}_t)$, producing a candidate-specific next state $s_{t+1}^{(k)}$. The SFT policy then continues greedily from that state until termination, subject to collector limits of \texttt{max\_steps}$=80$ and \texttt{max\_turn\_steps}$=20$, yielding a completed sibling trajectory $\tau_t^{(k)}$.

Because stochasticity is introduced only at the fork decision, sibling continuations compare alternative decisions under the same deterministic continuation policy. The resulting sibling trajectories expose the downstream outcomes of alternative decisions made from the same executable context.

\textbf{Fixed fork schedule and trunk reuse.}
Forks are selected only along the trunk; states reached inside sibling continuations do not become new fork points. The completed trunk provides the outcome for its retained decision at every fork. For at most $B$ forks and $K$ candidates per fork, including the trunk decision, the single-continuation schedule therefore requires at most
\[
N_{\mathrm{paths}}
\leq
1+B(K-1)
\]
completed paths: one shared trunk and $K-1$ sibling continuations per fork. Under the primary setting $B=10$ and $K=3$, this gives at most 21 completed paths per scenario. Execution and token cost additionally depend on continuation length. Unlike adaptive tree-search procedures such as MCTS-DPO~\citep{xie2024mctsdpo}, this collector uses completed outcomes only to compare decisions at the recorded fork states.

\subsection{Final Outcome Evaluation}

Each completed sibling first passes runtime and tool-execution checks. Its scenario-specific verifier is then evaluated on the completed continuation and terminal state. Let
\[
\mathcal{V}=\{v_1,\ldots,v_J\}
\]
denote the requirements associated with the scenario. These predicates encode applicable state, execution, and policy conditions. For a valid sibling terminating in $s_H^{(k)}$, the verified outcome is
\[
r_t^{(k)}
=
\prod_{j=1}^{J}
\mathbb{I}
\left[
v_j
\left(
\tau_t^{(k)},
s_H^{(k)}
\right)
=
1
\right].
\]

For each evaluated sibling, the collector retains its complete continuation, terminal state, predicate outcomes, execution metadata, and verified outcome.

\subsection{Preference-Pair Extraction}

Candidate assistant turns are canonicalized before comparison to remove duplicate decisions caused by transient identifiers. For the primary preference corpus, we require a minimum reward gap of $1.0$. Because the executable verifier produces binary outcomes, this retains only comparisons between a fully successful candidate and an unsuccessful candidate. For each fork, we retain at most one such pair satisfying
\[
r_t^+=1,
\qquad
r_t^-=0.
\]

The resulting preference example is
\[
x_t=(q,m_{<t},\mathcal{T}_t),
\qquad
y^+=u_t^+,
\qquad
y^-=u_t^-.
\]

The complete sibling continuations determine the preference label, while the divergent assistant turns serve as the preferred and rejected responses. Each pair is stored with its sibling continuations, terminal outcomes, and verification records as construction provenance. These pairs form $\mathcal{D}_{\mathrm{pref}}$.

\subsection{Preference-Data Distribution}

The preference corpus contains $1{,}150$ decision-level pairs with $1{,}049$ distinct serialized interaction prefixes. Preferred and rejected responses together contain 439 distinct serialized tool names in their tool-call blocks. Using the same tokenizer as Section~\ref{sec:corpus-statistics}, prefixes average $3{,}339.8$ tokens (range 40--$11{,}602$), while preferred and rejected responses average $68.4$ (11--669) and $69.9$ (10--647) tokens, respectively. Preferred and rejected responses contain an average of 1.12 and 1.20 explicit tool calls. These statistics reflect the intended supervision granularity: relatively short alternative decisions are compared within substantially longer interaction contexts. The $1{,}150$ preference pairs are constructed after SFT and are separate from the $30{,}395$ demonstration trajectories used for supervised fine-tuning.

\section{Training Objectives}
\label{app:training-details}

Davinci-Tool-30B is initialized from Qwen3-Coder-30B-A3B-Instruct~\citep{qwen2025qwen3}. All SFT demonstrations use the unified interaction serialization introduced in Section~\ref{sec:data-engine}.

\subsection{Training Configuration}
\label{app:training-config}

Table~\ref{tab:training-hyperparameters} summarizes the optimization settings for SFT and for DPO on the $1{,}150$-pair preference corpus. The SFT settings are shared by Davinci-Tool-30B-SFT, the Qwen3-8B and Qwen3-14B models trained on Davinci-ToolData, and the Qwen3-Coder-30B model trained on the public Toucan-1.5M release in Table~\ref{tab:main}. These runs use the same SFT hyperparameters, including four epochs, a learning rate of $1\times 10^{-5}$, and a maximum training sequence length of $32{,}768$ tokens; serialized examples exceeding this limit are truncated by the default SFT training pipeline. Their backbones and training corpora vary as specified in Section~\ref{sec:experiments}. SFT trains assistant tokens using the tokenizer's chat template. DPO uses precomputed reference log probabilities and keeps preference examples unpacked. Both stages use fused AdamW~\citep{loshchilov2019decoupled} and enable gradient checkpointing~\citep{chen2016sublinear}, FlashAttention~\citep{dao2022flashattention}, BF16, and DeepSpeed ZeRO stage 3~\citep{rajbhandari2020zero}, with no held-out validation split.

\begin{table}[htbp]
\centering
\small
\caption{Training hyperparameters. The SFT column applies to the 30B full-corpus model and the additional 8B, 14B, and Toucan-1.5M SFT runs. The DPO column applies to Davinci-Tool-30B. Batch size is per device, and gradient accumulation is measured in optimization steps.}
\label{tab:training-hyperparameters}
\begin{tabular}{@{}lcc@{}}
\toprule
\textbf{Setting}
& \textbf{SFT (shared settings)}
& \textbf{DPO} \\
\midrule
Optimizer & Fused AdamW & Fused AdamW \\
Learning rate & $1\times 10^{-5}$ & $5\times 10^{-6}$ \\
Epochs & 4 & 1 \\
Learning-rate schedule & Cosine & Cosine \\
Warmup fraction & 0 & 0.03 \\
Weight decay & 0.01 & 0.01 \\
Maximum sequence length & $32{,}768$ & 24,576 \\
Batch size per device & 1 & 1 \\
Gradient accumulation & 4 & 4 \\
Sample packing & Enabled & Disabled \\
Padding to maximum length & Enabled & Disabled \\
Precision & BF16 & BF16 \\
Random seed & 42 & 42 \\
Reference log probabilities & --- & Precomputed \\
Base preference scale & --- & 1.0 \\
\bottomrule
\end{tabular}
\end{table}

\subsection{Context Lengths and Evaluation Coverage}
\label{app:context-coverage}

The official Qwen3-8B and Qwen3-14B~\citep{qwen2025qwen3} model cards distinguish a native context length of $32{,}768$ tokens from an extended length of 131,072 tokens with YaRN~\citep{peng2024yarn}. Their released configurations set \texttt{max\_position\_embeddings} to $40{,}960$ with no RoPE scaling configured, whereas Qwen3-Coder-30B-A3B-Instruct is released with a 262,144-token context configuration~\citep{qwen2025qwen3}. We retain the default Qwen3-8B and Qwen3-14B positional configurations during fine-tuning and evaluation, with YaRN disabled. Both the original and SFT checkpoints use a $40{,}960$-token serving limit for the total input and generated output. This evaluation limit is distinct from the $32{,}768$-token maximum training sequence length shared by our SFT runs in Table~\ref{tab:training-hyperparameters}.

MCP-Atlas evaluates multi-step tool use from a user request, with subsequent calls returning observations that contribute to the interaction history~\citep{bandi2026mcpatlas}. The context budget must accommodate the prompt, visible tool schemas, retained assistant and tool messages, and generated output; the required length therefore depends on the realized trajectory. In our evaluation, accumulated interactions exceeded the $40{,}960$-token limit and prevented valid MCP-Atlas scores for Qwen3-8B and Qwen3-14B, both before and after SFT. We report these entries as dashes in Table~\ref{tab:main}; the SFT runs do not change this serving limit.

\subsection{Supervised Fine-Tuning}

The SFT training corpus is
\[
\mathcal{D}_{\mathrm{SFT}}
=
\mathcal{D}_{\mathrm{EGTS}}
\cup
\mathcal{D}_{\mathrm{EES\text{-}single}}
\cup
\mathcal{D}_{\mathrm{EES\text{-}cross}},
\]
with $13{,}824$ EGTS trajectories, $8{,}026$ EES-Single trajectories, and $8{,}545$ EES-Cross trajectories, for a total of $30{,}395$ demonstrations.

Each instance contains the system context, user task, visible tool schemas, assistant messages, executed tool calls, environment observations, and final response. Let $z$ denote the full serialized interaction and $\mathcal{A}(z)$ its assistant-token positions. SFT minimizes
\begin{equation}
\mathcal{L}_{\mathrm{SFT}}(\theta)
=
-\mathbb{E}_{z\sim\mathcal{D}_{\mathrm{SFT}}}
\left[
\sum_{i\in\mathcal{A}(z)}
\log\pi_\theta(z_i\mid z_{<i})
\right].
\label{eq:sft}
\end{equation}
Tool outputs remain in the conditioning context for subsequent assistant tokens but are excluded from the supervised loss.

Section~\ref{sec:ablations} compares six evaluated combinations of the three construction routes. The full SFT model uses all three.

\subsection{Direct Preference Optimization}

After SFT, the resulting checkpoint is used to construct $\mathcal{D}_{\mathrm{pref}}$ from restorable EES-Single and EES-Cross scenarios following Appendix~\ref{app:dpo-details}. The resulting preference corpus contains $1{,}150$ decision-level pairs. A frozen copy of the same SFT checkpoint serves as the reference policy $\pi_{\mathrm{ref}}$ for Direct Preference Optimization~\citep{rafailov2023dpo}.

For preference tuples $(x,y^+,y^-)$, DPO minimizes
\[
\mathcal{L}_{\mathrm{DPO}}(\theta)
=
-
\mathbb{E}_{(x,y^+,y^-)\sim\mathcal{D}_{\mathrm{pref}}}
\log
\sigma
\left(
\beta
\log
\frac{\pi_\theta(y^+\mid x)}
     {\pi_{\mathrm{ref}}(y^+\mid x)}
-
\beta
\log
\frac{\pi_\theta(y^-\mid x)}
     {\pi_{\mathrm{ref}}(y^-\mid x)}
\right).
\]

Only the divergent preferred and rejected assistant turns enter the response loss. Their complete downstream continuations are used to establish the outcome-based preference labels. The resulting checkpoint is reported as Davinci-Tool-30B in Section~\ref{sec:experiments}.

\section{Additional Related Work}
\label{app:additional-related}

Toolformer constructs supervision through tool-call augmentation~\citep{schick2023toolformer}, while MCP-Flow and TOUCAN scale trajectory collection across real-world MCP tools~\citep{wang2025mcpflow,xu2025toucan}. For executable environment synthesis, AutoWebWorld generates verifiable web environments from finite-state models~\citep{wu2026autowebworld}, while PROVE grounds task synthesis and reward signals in live MCP servers~\citep{abdelaziz2026prove}. These approaches are adjacent to the execution-first trajectory and domain-first environment methods discussed in Section~\ref{sec:related}. Our construction pipeline connects task-conditioned replay in authored systems, reusable executable-domain composition, and same-context preference construction within a unified training workflow.

%% file: references.bib
@article{schick2023toolformer,
  title={Toolformer: Language models can teach themselves to use tools},
  author={Schick, Timo and Dwivedi-Yu, Jane and Dess{\`\i}, Roberto and Raileanu, Roberta and Lomeli, Maria and Hambro, Eric and Zettlemoyer, Luke and Cancedda, Nicola and Scialom, Thomas},
  journal={Advances in neural information processing systems},
  volume={36},
  pages={68539--68551},
  year={2023}
}

@inproceedings{qin2023toolllm,
  title={Toolllm: Facilitating large language models to master 16000+ real-world apis},
  author={Qin, Yujia and Liang, Shihao and Ye, Yining and Zhu, Kunlun and Yan, Lan and Lu, Yaxi and Lin, Yankai and Cong, Xin and Tang, Xiangru and Qian, Bill and others},
  booktitle={International Conference on Learning Representations},
  volume={2024},
  pages={9695--9717},
  year={2024}
}

@inproceedings{li2023apibank,
  title={Api-bank: A comprehensive benchmark for tool-augmented llms},
  author={Li, Minghao and Zhao, Yingxiu and Yu, Bowen and Song, Feifan and Li, Hangyu and Yu, Haiyang and Li, Zhoujun and Huang, Fei and Li, Yongbin},
  booktitle={Proceedings of the 2023 conference on empirical methods in natural language processing},
  pages={3102--3116},
  year={2023}
}

@inproceedings{liu2024toolace,
  title={Toolace: Winning the points of llm function calling},
  author={Liu, Weiwen and Huang, Xu and Zeng, Xingshan and Yu, Shuai and Li, Dexun and Wang, Shuai and Gan, Weinan and Liu, Zhengying and Yu, Yuanqing and WANG, Zezhong and others},
  booktitle={International conference on learning representations},
  volume={2025},
  pages={41359--41381},
  year={2025}
}

@inproceedings{zhou2023webarena,
  title={Webarena: A realistic web environment for building autonomous agents},
  author={Zhou, Shuyan and Xu, Frank F and Zhu, Hao and Zhou, Xuhui and Lo, Robert and Sridhar, Abishek and Cheng, Xianyi and Ou, Tianyue and Bisk, Yonatan and Fried, Daniel and others},
  booktitle={International Conference on Learning Representations},
  volume={2024},
  pages={15585--15606},
  year={2024}
}

@article{yao2024taubench,
  title={{$\tau$}-bench: A Benchmark for Tool-Agent-User Interaction in Real-World Domains},
  author={Yao, Shunyu and Shinn, Noah and Razavi, Pedram and Narasimhan, Karthik},
  journal={arXiv preprint arXiv:2406.12045},
  year={2024}
}

@article{barres2025tau2bench,
  title={{$\tau^{2}$}-Bench: Evaluating Conversational Agents in a Dual-Control Environment},
  author={Barres, Victor and Dong, Honghua and Ray, Soham and Si, Xujie and Narasimhan, Karthik},
  journal={arXiv preprint arXiv:2506.07982},
  year={2025}
}

@inproceedings{patil2025bfcl,
  title={The berkeley function calling leaderboard (bfcl): From tool use to agentic evaluation of large language models},
  author={Patil, Shishir G and Mao, Huanzhi and Yan, Fanjia and Ji, Charlie Cheng-Jie and Suresh, Vishnu and Stoica, Ion and Gonzalez, Joseph E},
  booktitle={Forty-second International Conference on Machine Learning},
  year={2025}
}

@article{bandi2026mcpatlas,
  title={Mcp-atlas: A large-scale benchmark for tool-use competency with real mcp servers},
  author={Bandi, Chaithanya and Dumitru, Razvan-Gabriel and Hertzberg, Ben and Agarwal, Divyansh and Boo, Geobio and Polakam, Tejas and Hassaan, Sami and Da, Jeff and Kim, HiJae and Gupta, Vipul and others},
  journal={arXiv preprint arXiv:2602.00933},
  year={2026}
}

@article{qwen2025qwen3,
  title={Qwen3 technical report},
  author={Yang, An and Li, Anfeng and Yang, Baosong and Zhang, Beichen and Hui, Binyuan and Zheng, Bo and Yu, Bowen and Gao, Chang and Huang, Chengen and Lv, Chenxu and others},
  journal={arXiv preprint arXiv:2505.09388},
  year={2025}
}

@inproceedings{rajbhandari2020zero,
  title={Zero: Memory optimizations toward training trillion parameter models},
  author={Rajbhandari, Samyam and Rasley, Jeff and Ruwase, Olatunji and He, Yuxiong},
  booktitle={SC20: international conference for high performance computing, networking, storage and analysis},
  pages={1--16},
  year={2020},
  organization={IEEE}
}

@article{loshchilov2019decoupled,
  title={Decoupled weight decay regularization},
  author={Loshchilov, Ilya and Hutter, Frank},
  journal={arXiv preprint arXiv:1711.05101},
  year={2017}
}

@inproceedings{lu2024toolsandbox,
  title={Toolsandbox: A stateful, conversational, interactive evaluation benchmark for llm tool use capabilities},
  author={Lu, Jiarui and Holleis, Thomas and Zhang, Yizhe and Aumayer, Bernhard and Nan, Feng and Bai, Haoping and Ma, Shuang and Ma, Shen and Li, Mengyu and Yin, Guoli and others},
  booktitle={Findings of the Association for Computational Linguistics: NAACL 2025},
  pages={1160--1183},
  year={2025}
}

@inproceedings{trivedi2024appworld,
  title={Appworld: A controllable world of apps and people for benchmarking interactive coding agents},
  author={Trivedi, Harsh and Khot, Tushar and Hartmann, Mareike and Manku, Ruskin and Dong, Vinty and Li, Edward and Gupta, Shashank and Sabharwal, Ashish and Balasubramanian, Niranjan},
  booktitle={Proceedings of the 62nd Annual Meeting of the Association for Computational Linguistics (Volume 1: Long Papers)},
  pages={16022--16076},
  year={2024}
}

@article{xie2024osworld,
  title={Osworld: Benchmarking multimodal agents for open-ended tasks in real computer environments},
  author={Xie, Tianbao and Zhang, Danyang and Chen, Jixuan and Li, Xiaochuan and Zhao, Siheng and Cao, Ruisheng and Hua, Toh J and Cheng, Zhoujun and Shin, Dongchan and Lei, Fangyu and others},
  journal={Advances in Neural Information Processing Systems},
  volume={37},
  pages={52040--52094},
  year={2024}
}

@article{yao2022webshop,
  title={Webshop: Towards scalable real-world web interaction with grounded language agents},
  author={Yao, Shunyu and Chen, Howard and Yang, John and Narasimhan, Karthik},
  journal={Advances in Neural Information Processing Systems},
  volume={35},
  pages={20744--20757},
  year={2022}
}

@article{yang2024sweagent,
  title={Swe-agent: Agent-computer interfaces enable automated software engineering},
  author={Yang, John and Jimenez, Carlos and Wettig, Alexander and Lieret, Kilian and Yao, Shunyu and Narasimhan, Karthik and Press, Ofir},
  journal={Advances in Neural Information Processing Systems},
  volume={37},
  pages={50528--50652},
  year={2024}
}

@inproceedings{zheng2025deepresearcher,
  title={Deepresearcher: Scaling deep research via reinforcement learning in real-world environments},
  author={Zheng, Yuxiang and Fu, Dayuan and Hu, Xiangkun and Cai, Xiaojie and Ye, Lyumanshan and Lu, Pengrui and Liu, Pengfei},
  booktitle={Proceedings of the 2025 Conference on Empirical Methods in Natural Language Processing},
  pages={414--431},
  year={2025}
}

@article{li2025webthinker,
  title={Webthinker: Empowering large reasoning models with deep research capability},
  author={Li, Xiaoxi and Jin, Jiajie and Dong, Guanting and Qian, Hongjin and Wu, Yongkang and Wen, Ji-Rong and Zhu, Yutao and Dou, Zhicheng},
  journal={Advances in Neural Information Processing Systems},
  volume={38},
  pages={120091--120131},
  year={2026}
}

@article{yang2025exif,
  title={Automated skill discovery for language agents through exploration and iterative feedback},
  author={Yang, Yongjin and Kang, Sinjae and Lee, Juyong and Lee, Dongjun and Yun, Se-Young and Lee, Kimin},
  journal={arXiv preprint arXiv:2506.04287},
  year={2025}
}

@inproceedings{ramrakhya2025autoplay,
  title={Scaling synthetic task generation for agents via exploration},
  author={Ramrakhya, Ram and Szot, Andrew and Attia, Omar and Mazoure, Bogdan and Nguyen, Anh and Yang, Yuhao and Gan, Zhe and Agrawal, Harsh and Toshev, Alexander},
  booktitle={International Conference on Learning Representations},
  volume={2026},
  pages={123630--123660},
  year={2026}
}

@article{xu2025toucan,
  title={Toucan: Synthesizing 1.5 m tool-agentic data from real-world mcp environments},
  author={Xu, Zhangchen and Soria, Adriana Meza and Tan, Shawn and Roy, Anurag and Agrawal, Ashish Sunil and Poovendran, Radha and Panda, Rameswar},
  journal={arXiv preprint arXiv:2510.01179},
  year={2025}
}

@inproceedings{wang2025mcpflow,
  title={Mcp-flow: Facilitating llm agents to master real-world, diverse and scaling mcp tools},
  author={Wang, Wenhao and Niu, Peizhi and Xu, Zhao and Chen, Zhaoyu and Du, Jian and Du, Yaxin and Pang, Xianghe and Huang, Keduan and Wang, Yanfeng and Yan, Qiang and others},
  booktitle={Proceedings of the 64th Annual Meeting of the Association for Computational Linguistics (Volume 1: Long Papers)},
  pages={5087--5116},
  year={2026}
}

@article{zhai2025agentevolver,
  title={Agentevolver: Towards efficient self-evolving agent system},
  author={Zhai, Yunpeng and Tao, Shuchang and Chen, Cheng and Zou, Anni and Chen, Ziqian and Fu, Qingxu and Mai, Shinji and Yu, Li and Deng, Jiaji and Cao, Zouying and others},
  journal={arXiv preprint arXiv:2511.10395},
  year={2025}
}

@inproceedings{wang2026trajectory2task,
  title={Trajectory2task: Training robust tool-calling agents with synthesized yet verifiable data for complex user intents},
  author={Wang, Ziyi and Lu, Yuxuan and Zhang, Yimeng and Chen, Pei and Dong, Ziwei and Huang, Jing and Gesi, Jiri and Tang, Xianfeng and Luo, Chen and Liu, Qun and others},
  booktitle={Proceedings of the 64th Annual Meeting of the Association for Computational Linguistics (Volume 1: Long Papers)},
  pages={44021--44044},
  year={2026}
}

@article{tian2026astra,
  title={Astra: Automated synthesis of agentic trajectories and reinforcement arenas},
  author={Tian, Xiaoyu and Wang, Haotian and Chen, Shuaiting and Zhou, Hao and Yu, Kaichi and Zhang, Yudian and Ouyang, Jade and Yin, Junxi and Chen, Jiong and Guo, Baoyan and others},
  journal={arXiv preprint arXiv:2601.21558},
  year={2026}
}

@article{wu2026autowebworld,
  title={Autowebworld: Synthesizing infinite verifiable web environments via finite state machines},
  author={Wu, Yifan and Peng, Yiran and Chen, Yiyu and Ruan, Jianhao and Zhuang, Zijie and Yang, Cheng and Zhang, Jiayi and Chen, Man and Tseng, Yenchi and Yu, Zhaoyang and others},
  journal={arXiv preprint arXiv:2602.14296},
  year={2026}
}

@article{rafailov2023dpo,
  title={Direct preference optimization: Your language model is secretly a reward model},
  author={Rafailov, Rafael and Sharma, Archit and Mitchell, Eric and Manning, Christopher D and Ermon, Stefano and Finn, Chelsea},
  journal={Advances in neural information processing systems},
  volume={36},
  pages={53728--53741},
  year={2023}
}

@inproceedings{song2024eto,
  title={Trial and error: Exploration-based trajectory optimization of LLM agents},
  author={Song, Yifan and Yin, Da and Yue, Xiang and Huang, Jie and Li, Sujian and Lin, Bill Yuchen},
  booktitle={Proceedings of the 62nd Annual Meeting of the Association for Computational Linguistics (Volume 1: Long Papers)},
  pages={7584--7600},
  year={2024}
}

@article{putta2024agentq,
  title={Agent q: Advanced reasoning and learning for autonomous ai agents},
  author={Putta, Pranav and Mills, Edmund and Garg, Naman and Motwani, Sumeet and Finn, Chelsea and Garg, Divyansh and Rafailov, Rafael},
  journal={arXiv preprint arXiv:2408.07199},
  year={2024}
}

@article{xie2024mctsdpo,
  title={Monte carlo tree search boosts reasoning via iterative preference learning},
  author={Xie, Yuxi and Goyal, Anirudh and Zheng, Wenyue and Kan, Min-Yen and Lillicrap, Timothy P and Kawaguchi, Kenji and Shieh, Michael},
  journal={arXiv preprint arXiv:2405.00451},
  year={2024}
}

@article{wang2024alphallmcpl,
  title={Towards self-improvement of llms via mcts: Leveraging stepwise knowledge with curriculum preference learning},
  author={Wang, Xiyao and Song, Linfeng and Tian, Ye and Yu, Dian and Peng, Baolin and Mi, Haitao and Huang, Furong and Yu, Dong},
  journal={arXiv preprint arXiv:2410.06508},
  year={2024}
}

@inproceedings{xu2025agenttrek,
  title={Agenttrek: Agent trajectory synthesis via guiding replay with web tutorials},
  author={Xu, Yiheng and Lu, Dunjie and Shen, Zhennan and Wang, Junli and Wang, Zekun and Mao, Yuchen and Xiong, Caiming and Yu, Tao},
  booktitle={International Conference on Learning Representations},
  volume={2025},
  pages={79822--79843},
  year={2025}
}

@inproceedings{song2026envscaler,
  title={Envscaler: Scaling tool-interactive environments for llm agent via programmatic synthesis},
  author={Song, Xiaoshuai and Chang, Haofei and Dong, Guanting and Zhu, Yutao and Wen, Ji-Rong and Dou, Zhicheng},
  booktitle={Findings of the Association for Computational Linguistics: ACL 2026},
  pages={8326--8357},
  year={2026}
}

@article{abdelaziz2026prove,
  title={Synthesize and Reward--Reinforcement Learning for Multi-Step Tool Use in Live Environments},
  author={Abdelaziz, Ibrahim and Munawar, Asim and Basu, Kinjal and Crouse, Maxwell and Gunasekara, Chulaka and Katrekar, Suneet and Kapanipathi, Pavan},
  journal={arXiv preprint arXiv:2606.03892},
  year={2026}
}

@article{ouajdi2026executionfirst,
  title={Execution-First Synthetic Tool-Use Trace Generation for LLM Agents},
  author={Ouajdi, Hafsa and Giannuzzo, Francesco and Boukhary, Alaa and Papotti, Paolo and Conangla, Gerard and Elwood, Adam},
  journal={arXiv preprint arXiv:2607.29175},
  year={2026}
}

@inproceedings{guo2024stabletoolbench,
  title={Stabletoolbench: Towards stable large-scale benchmarking on tool learning of large language models},
  author={Guo, Zhicheng and Cheng, Sijie and Wang, Hao and Liang, Shihao and Qin, Yujia and Li, Peng and Liu, Zhiyuan and Sun, Maosong and Liu, Yang},
  booktitle={Findings of the Association for Computational Linguistics: ACL 2024},
  pages={11143--11156},
  year={2024}
}

@inproceedings{wu2025mcpmark,
  title={Mcpmark: A benchmark for stress-testing realistic and comprehensive mcp use},
  author={Wu, Zijian and Liu, Xiangyan and Chen, Lingjun and Meng, Fanqing and Du, Lingxiao and Zhao, Yiran and Zhang, Fanshi and Ye, Yaoqi and Wang, Jiawei and Wang, Zirui and others},
  booktitle={International Conference on Learning Representations},
  volume={2026},
  pages={79852--79895},
  year={2026}
}

@inproceedings{merrill2026terminalbench,
  title={Terminal-bench: Benchmarking agents on hard, realistic tasks in command line interfaces},
  author={Merrill, Mike and Shaw, Alexander and Carlini, Nicholas and Li, Boxuan and Raj, Harsh and Bercovich, Ivan and Shi, Lin and Shin, Jeong and Walshe, Thomas and Buchanan, E Kelly and others},
  booktitle={International Conference on Learning Representations},
  volume={2026},
  pages={40903--40986},
  year={2026}
}

@inproceedings{shi2017worldofbits,
  title={World of bits: An open-domain platform for web-based agents},
  author={Shi, Tianlin and Karpathy, Andrej and Fan, Linxi and Hernandez, Jonathan and Liang, Percy},
  booktitle={International Conference on Machine Learning},
  pages={3135--3144},
  year={2017},
  organization={PMLR}
}

@inproceedings{wang2022scienceworld,
  title={Scienceworld: Is your agent smarter than a 5th grader?},
  author={Wang, Ruoyao and Jansen, Peter and C{\^o}t{\'e}, Marc-Alexandre and Ammanabrolu, Prithviraj},
  booktitle={Proceedings of the 2022 Conference on Empirical Methods in Natural Language Processing},
  pages={11279--11298},
  year={2022}
}

@article{jansen2024discoveryworld,
  title={Discoveryworld: A virtual environment for developing and evaluating automated scientific discovery agents},
  author={Jansen, Peter and C{\^o}t{\'e}, Marc-Alexandre and Khot, Tushar and Bransom, Erin and Dalvi Mishra, Bhavana and Majumder, Bodhisattwa Prasad and Tafjord, Oyvind and Clark, Peter},
  journal={Advances in neural information processing systems},
  volume={37},
  pages={10088--10116},
  year={2024}
}

@inproceedings{li2026toolathlon,
  title={The tool decathlon: Benchmarking language agents for diverse, realistic, and long-horizon task execution},
  author={Li, Junlong and Zhao, Wenshuo and Zhao, Jian and Zeng, Weihao and Wu, Haoze and Wang, Xiaochen and Ge, Rui and Cao, Yuxuan and Huang, Yuzhen and Liu, Wei and others},
  booktitle={International Conference on Learning Representations},
  volume={2026},
  pages={53359--53394},
  year={2026}
}

@article{kimi2025k2,
  title={Kimi k2: Open agentic intelligence},
  author={Team, Kimi and Bai, Yifan and Bao, Yiping and Charles, Y and Chen, Cheng and Chen, Guanduo and Chen, Haiting and Chen, Huarong and Chen, Jiahao and Chen, Ningxin and others},
  journal={arXiv preprint arXiv:2507.20534},
  year={2025}
}

@article{chen2016sublinear,
  title={Training deep nets with sublinear memory cost},
  author={Chen, Tianqi and Xu, Bing and Zhang, Chiyuan and Guestrin, Carlos},
  journal={arXiv preprint arXiv:1604.06174},
  year={2016}
}

@misc{minimax2025m21,
  author       = {{MiniMax}},
  title        = {{MiniMax M2.1}: Significantly Enhanced Multi-Language Programming, Built for Real-World Complex Tasks},
  year         = {2025},
  howpublished = {\url{https://www.minimax.io/news/minimax-m21}}
}

@misc{zai2025glm47,
  author       = {{Z.ai}},
  title        = {{GLM-4.7}},
  year         = {2025},
  howpublished = {\url{https://z.ai/blog/glm-4.7}}
}

@misc{openai2025gpt52,
  author       = {{OpenAI}},
  title        = {Introducing {GPT-5.2}},
  year         = {2025},
  howpublished = {\url{https://openai.com/index/introducing-gpt-5-2/}}
}

@misc{anthropic2025claudesonnet45,
  author       = {{Anthropic}},
  title        = {Introducing {Claude Sonnet 4.5}},
  year         = {2025},
  howpublished = {\url{https://www.anthropic.com/news/claude-sonnet-4-5}}
}

@article{deepseekai2026deepseekv4,
  title={Deepseek-v4: Towards highly efficient million-token context intelligence},
  author={Xu, Anyi and Lin, Bangcai and Xue, Bing and Wang, Bingxuan and Xu, Bingzheng and Wu, Bochao and Zhang, Bowei and Lin, Chaofan and Dong, Chen and Ling, Chenchen and others},
  journal={arXiv preprint arXiv:2606.19348},
  year={2026}
}

@inproceedings{peng2024yarn,
  title={Yarn: Efficient context window extension of large language models},
  author={Peng, Bowen and Quesnelle, Jeffrey and Fan, Honglu and Shippole, Enrico},
  booktitle={International Conference on Learning Representations},
  volume={2024},
  pages={31932--31951},
  year={2024}
}

@article{dao2022flashattention,
  title={Flashattention: Fast and memory-efficient exact attention with io-awareness},
  author={Dao, Tri and Fu, Dan and Ermon, Stefano and Rudra, Atri and R{\'e}, Christopher},
  journal={Advances in neural information processing systems},
  volume={35},
  pages={16344--16359},
  year={2022}
}
